\documentclass[10pt,twocolumn,letterpaper]{article}

\usepackage[pagenumbers]{cvpr} 
\usepackage{mathtools}
\usepackage[dvipsnames]{xcolor}

\usepackage{times}
\usepackage{epsfig}
\usepackage{graphicx}
\usepackage{amsmath}
\usepackage{amssymb}
\usepackage{caption}
\usepackage{subcaption}
\usepackage{interval}
\usepackage{booktabs}
\usepackage{multirow}
\usepackage{textcmds}

\usepackage{algorithm,algpseudocode}

\usepackage{colortbl}
\usepackage{tabulary}
\usepackage{etoolbox}
\usepackage[dvipsnames]{xcolor}

\definecolor{tabfirst}{rgb}{1, 0.7, 0.7} 
\definecolor{tabsecond}{rgb}{1, 0.85, 0.7}
\definecolor{tabthird}{rgb}{1, 1, 0.7} 
\definecolor{cvprblue}{rgb}{0.21,0.49,0.74}
\usepackage[pagebackref,breaklinks,colorlinks,citecolor=cvprblue]{hyperref}

\def\confName{CVPR}
\def\confYear{2024}

\title{Integrating Efficient Optimal Transport and Functional Maps For Unsupervised Shape Correspondence Learning}

\author{Tung Le$^{1}$ $\qquad$ Khai Nguyen$^2$ $\qquad$ Shanlin Sun$^1$ $\qquad$ Nhat Ho$^2$ $\qquad$ Xiaohui Xie$^1$\\
$^1$University of California, Irvine \\
$^2$The University of Texas at Austin \\
}

\begin{document}
\maketitle
\begin{abstract}

In the realm of computer vision and graphics, accurately establishing correspondences between geometric 3D shapes is pivotal for applications like object tracking, registration, texture transfer, and statistical shape analysis. Moving beyond traditional hand-crafted and data-driven feature learning methods, we incorporate spectral methods with deep learning, focusing on functional maps (FMs) and optimal transport (OT). Traditional OT-based approaches, often reliant on entropy regularization OT in learning-based framework, face computational challenges due to their quadratic cost. Our key contribution is to employ the sliced Wasserstein distance (SWD) for OT, which is a valid fast optimal transport metric in an unsupervised shape matching framework. This unsupervised framework integrates functional map regularizers with a novel OT-based loss derived from SWD, enhancing feature alignment between shapes treated as discrete probability measures. We also introduce an adaptive refinement process utilizing entropy regularized OT, further refining feature alignments for accurate point-to-point correspondences. Our method demonstrates superior performance in non-rigid shape matching, including near-isometric and non-isometric scenarios, and excels in downstream tasks like segmentation transfer. The empirical results on diverse datasets highlight our framework's effectiveness and generalization capabilities, setting new standards in non-rigid shape matching with efficient OT metrics and an adaptive refinement module.  
\vspace{-1em}
\end{abstract}    
\section{Introduction}
\label{sec:intro}

Establishing precise correspondences between geometric 3D shapes is a core challenge in various domains of computer vision and graphics, including but not limited to, object tracking, registration, reconstruction, deformation, texture transfer, and statistical shape analysis~\cite{zhou2016fast, dinh2005texture, bogo2014faust, sumner2004deformation, sun2023hybrid, han2024hybrid}. To facilitate the mapping between non-rigid shapes, early approaches~\cite{bronstein2010scale, salti2014shot, aubry2011wave} concentrated on the development of hand-crafted features, leveraging geometric invariance as a key principle. In the latter approaches~\cite{attaiki2023ncp, donati2022deep, cao2022unsupervised, li2022learning}, there has been a shift towards the utilization of data-driven methods for feature learning, which has resulted in marked enhancements in terms of accuracy, efficiency, and robustness.

Recently, an increasing body of work has exploited the use of spectral methods~\cite{litany2017deep, halimi2019unsupervised, roufosse2019unsupervised, attaiki2021dpfm, eisenberger2020deep}, especially the functional map (FM) representation~\cite{ovsjanikov2012functional}. Specifically, the FM methods succinctly encode correspondences through compact matrices, utilizing a truncated spectral basis. With recent developments in deep learning, deep FM (DFM) is quickly employed in numerous settings~\cite{sun2023spatially, li2022learning, cao2023unsupervised, cao2023self} by incorporating feature learning as geometric descriptors for FM frameworks. Most DFM works focus on learning features that optimize FM priors to express desirable map priors, e.g. area preservation, isometry, and bijectivity, which achieves remarkable results even without supervision~\cite{halimi2019unsupervised, roufosse2019unsupervised, ginzburg2020cyclic, cao2022unsupervised, cao2023unsupervised}. On the other hand, less attention is paid to the problem of explicitly aligning features outputted from the feature extractor network, due to the lack of smoothness and consistency of linear assignment problems. 

In this work, we focus on jointly learning features via the functional map, and explicit features, i.e. directly from the feature extractor to establish correct correspondence. Nonetheless, learning to map explicit features is not easy since the geometric objects might potentially undergo arbitrary deformations. Therefore, we propose to employ optimal transport (OT), which is a well-known approach for linear assignment problems, to cast the feature alignment from 3D shapes as a probability measures matching problem. 

The Wasserstein distance~\cite{peyre2019computational, villani2009optimal} is widely acknowledged as an effective OT metric for comparing two probability measures, particularly when their supports are disjoint. However, it comes with the drawback of high computational complexity. Specifically, for discrete probability measures with at most $m$ supports, the time and memory complexities are $\mathcal{O}(m^3 \log m)$ and $\mathcal{O}(m^2)$, respectively. This computational burden is exacerbated in 3D shape applications where each shape, represented as mesh, is treated as a distinct probability measure. To ameliorate the computational demands, entropic regularization coupled with the Sinkhorn algorithm~\cite{cuturi2013sinkhorn} can yield an $\epsilon$-approximation of the Wasserstein distance with a time complexity of $\mathcal{O}(m^2/\epsilon^2)$~\cite{altschuler2017near, lin2019efficient, Lin-2019-Acceleration, Lin-2020-Revisiting}. Nonetheless, this method does not alleviate the $\mathcal{O}(m^2)$ memory complexity due to the necessity of storing the cost matrix. Additionally, the entropic regularization fails to produce a valid metric between probability measures as it does not satisfy the triangle inequality. An alternative, more efficient method is the sliced Wasserstein distance (SWD)~\cite{bonneel2015sliced}, which calculates the expectation of the Wasserstein distance between random one-dimensional push-forward measures derived from the original measures. SWD offers a time complexity of $\mathcal{O}(m \log m)$ and a linear memory complexity of $\mathcal{O}(m)$.

Motivated by the above discussion, we introduce a novel differentiable unsupervised OT-based loss derived from efficient sliced Wasserstein distance, which accounts for associating two extracted extrinsic features to align two meshes combined with functional map regularizers. Our proposed approach leverages a valid efficient OT metric to obtain highly discriminative local feature matching. Additionally, the integration of functional map regularizers promotes smoothness in the mapping process, allowing our method to achieve both precise and smooth correspondence.

Furthermore, we introduce an adaptive refinement process tailored for each pair of shapes, utilizing entropy regularized OT to enhance matching performance. The differentiable nature of entropic regularization in OT enables our refinement strategy to leverage the Sinkhorn algorithm. This approach yields a soft point-wise map, which is instrumental in calculating FM regularizers. These regularizers are then used to iteratively update features, thereby facilitating the retrieval of precise point-to-point correspondences.

Finally, we demonstrate our proposed approach on a diverse and extensive selection of datasets. Our contributions are as follows: 
\begin{itemize}
    \item We propose an unsupervised learning framework that employs efficient optimal transport to jointly learn with functional map in shape matching paradigm. Subsequently, we derive two novel unsupervised loss functions based on sliced Wasserstein distance, which is a valid fast optimal transport metric, to effectively align mesh features by interpreting them as probability measures, potentially offering a promising avenue for advancements in shape matching through efficient optimal transport.
    \item To enhance the quality of point mapping,  we propose an adaptive refinement module that iteratively refines the optimal transport similarity matrix estimated via entropy regularization optimal transport.
    \item We outperform previous state-of-the-art works in various settings of non-rigid shape matching including near-isometric and non-isometric shape matching. Additionally, when applied to a downstream task such as segmentation transfer, our approach continues to outperform contemporary state-of-the-art methods in non-rigid shape matching. This success not only demonstrates the efficacy of our method in specific applications but also underlines its strong generalization capabilities across various use cases in shape matching. 
\end{itemize}

\section{Related work}
\label{sec:related-work}
Shape matching has been extensively explored for decades. For a comprehensive examination of this topic, we encourage readers to consult the detailed analyses presented in surveys~\cite{sahilliouglu2020recent, van2011survey}. In this section, we focus specifically on the literature subset that directly relates to our research objectives.

\subsection{Deep functional maps for shape correspondence.}
Our methodology is founded on the functional map representation, initially introduced in~\cite{ovsjanikov2012functional} and substantially developed through subsequent research, e.g.~\cite{ovsjanikov2016computing}. The central concept of functional maps revolves around expressing shape correspondences as transformations between their respective spectral embeddings. This is efficiently achieved by utilizing compact matrices formulated from reduced eigenbases. The functional maps approach has seen considerable enhancements in terms of accuracy, efficiency, and robustness, as evidenced by a variety of recent contributions~\cite{kovnatsky2013coupled, huang2014functional, rodola2017partial}. In contrast to axiomatic approaches that rely on manually engineered features~\cite{sun2009concise}, deep functional map methods aim to autonomously learn features from training data. The pioneering work in this domain was FMNet~\cite{litany2017deep}, which introduced a method to learn non-linear transformations of SHOT descriptors~\cite{salti2014shot}. Subsequent developments~\cite{halimi2019unsupervised, roufosse2019unsupervised} facilitated the unsupervised training of FMNet by incorporating isometry losses in both spatial and spectral domains. This unsupervised approach has been further enhanced with the advent of robust mesh feature extractors~\cite{sharp2020diffusionnet}, leading to the development of new frameworks~\cite{cao2023unsupervised, li2022learning, cao2022unsupervised, donati2022deep}  that learn directly from geometric data, achieving top-tier performance.

\subsection{Optimal transport for shape correspondence}
Optimal transport has emerged as a powerful tool in the field of shape correspondence, offering innovative approaches to match and analyze complex shapes in computer graphics and computer vision. Starting with the axiomatic shape matching approach, \cite{solomon2016entropic} proposed an algorithm for probabilistic correspondence that optimizes an entropy-regularized Gromov-Wasserstein (GW) objective~\cite{memoli2011gromov} to find the correspondence between two given shapes. The proposed framework is inefficient since solving entropy-regularized GW objective is relatively expensive and it does not perform well on non-isometric shape matching. To address the computational overhead of solving OT cost, \cite{shen2021accurate} brought robust OT to the forefront, significantly enhancing the accuracy and efficiency of point cloud registration, but the framework is designed for point cloud that avoids the connectivity of the shape mesh. Perhaps the most relevant work to ours is Deep Shells~\cite{eisenberger2020deep}, which is an improvement of~\cite{eisenberger2020smooth}. Deep Shells demonstrated how OT can be seamlessly integrated into deep neural networks, offering a new perspective in shape matching with improved adaptability and precision. However, computing OT cost via Sinkhorn algorithm in Deel Shells~\cite{eisenberger2020deep} can be expensive since it has to store the cost matrix with quadratic memory cost and quadratic time complexity. In light of this, we propose to employ an efficient OT in learning shape correspondence. To be specific, we employ sliced Wasserstein distance, which calculates the expectation the Wasserstein distance between two random one-dimensional push-forward measures derived from original measures. Recently, sliced Wasserstein distance has been successfully applied in point cloud~\cite{nguyen2021point} and shape~\cite{le2023diffeomorphic} deformation. However, to the best of our knowledge, we are the first to employ sliced Wasserstein distance on shape correspondence framework.

\section{Background}
\label{sec:background}
In this section, we briefly recap functional map representation~\cite{ovsjanikov2012functional}. After that, we review the definition of Wasserstein distance and its closed-formed solution sliced Wasserstein distance.

\begin{figure*}[t]
  \begin{center}
    \includegraphics[width=\textwidth]{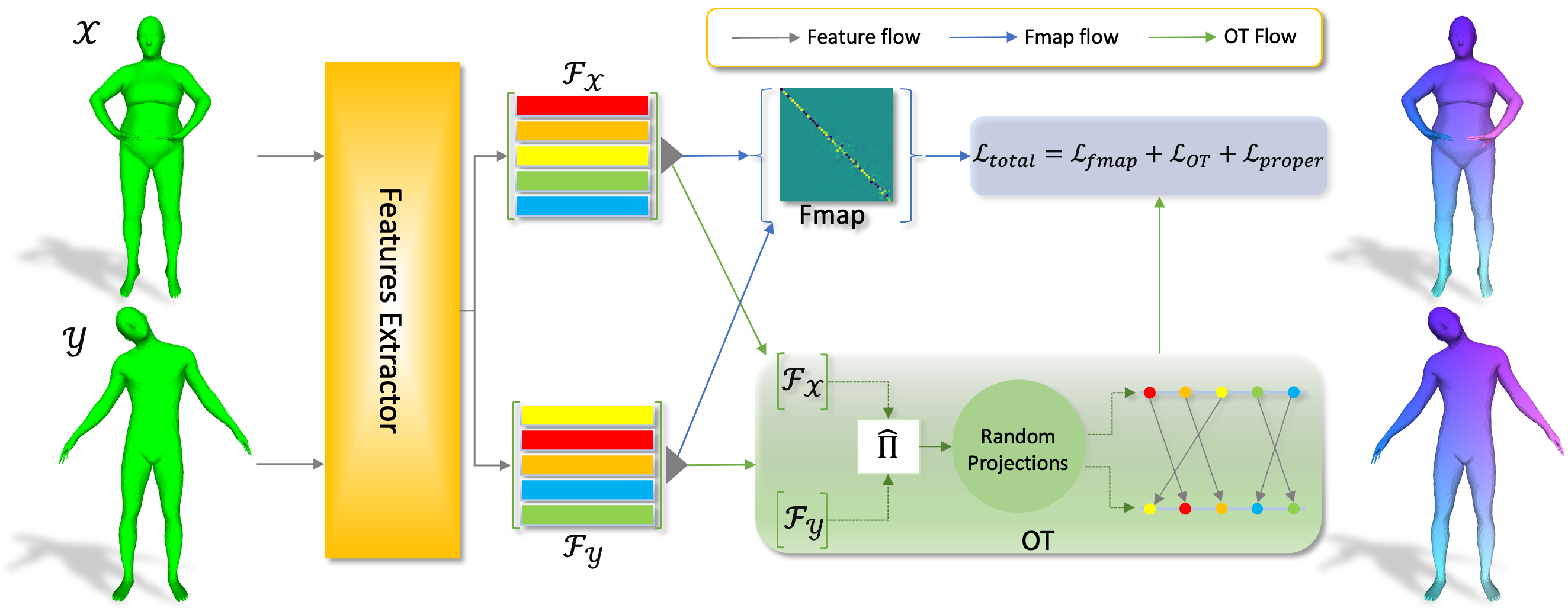}
  \end{center}
  \caption{\textbf{Overview of unsupervised shape matching via efficient OT.} Our framework takes as input a pair of shapes $\mathcal{X}$ and $\mathcal{Y}$ and outputs point-to-point correspondence. Firstly, the features extractor tasks the pair input and extracts vertex-wise features $\mathcal{F}_x$ and $\mathcal{F}_y$. Subsequently, the differentiable functional map solver is used to compute functional map given pre-computed eigenfunctions and the extracted features. In parallel, our framework estimates a soft feature similarity matrix, derived from the same extracted features. After that, an OT cost is computed given soft feature similarity and extracted feature $\mathcal{F}_x$ and $\mathcal{F}_y$. Finally, a proper loss is optimized together with regularized functional map loss and OT loss.}
  \label{fig:framework}
  \vspace{-0.8em}
\end{figure*}

\subsection{(Deep) Functional Maps}
\label{subsec:FM}
Given a pair of smooth shapes $\mathcal{X}$ and $\mathcal{Y}$, which are discretized as triangular meshes with $n_x$ and $n_y$ vertices, respectively. The functional map method aims to obtain a dense correspondence between the two shapes by compactly representing the correspondence matrix as a smaller matrix. Specifically, the leading $k$ eigenfunctions of the Laplace-Beltrami operator are computed on both shapes $\mathcal{X}$, $\mathcal{Y}$ and are presented as $\Phi_x \in \mathbb{R}^{n_x \times k}$ and $\Phi_y \in \mathbb{R}^{n_y \times k}$, respectively. The geometric features of the shape are either precomputed~\cite{salti2014shot} or extracted from a neural network~\cite{sharp2020diffusionnet}, represented as $\mathcal{F}_x \in \mathbb{R}^{n_x \times d}$ and $\mathcal{F}_y \in \mathbb{R}^{n_y \times d}$, where $d$ is the feature dimension. The extracted features are then projected into the eigenbasis to get the corresponding coefficients $\mathbf{A} = \Phi_x^{\dagger} \mathcal{F}_x \in \mathbb{R}^{k \times d}$ and $\mathbf{B} = \Phi_y^{\dagger} \mathcal{F}_y \in \mathbb{R}^{k \times d}$, where $\dagger$ denotes the Moore-Penrose pseudo-inverse. After that, the bidirectional optimal functional map $\mathbf{C}_{xy}^{\ast}, \mathbf{C}_{yx}^{\ast} \in \mathbb{R}^{k \times k}$ is obtained by solving the linear system:
\begin{equation}
    \label{eq:fmap}
    \mathbf{C}_{xy}^{\ast} = \arg \min_{\mathbf{C}} E_{data}(\mathbf{C}) + E_{reg}(\mathbf{C}),
\end{equation}
where $E_{data}(\mathbf{C}) = \| \mathbf{CA} - \mathbf{B} \|^2$ promotes the descriptor preservation, whereas the $E_{reg}$ is a regularization term imposing structural properties of $\mathbf{C}$~\cite{ovsjanikov2012functional}. Finally, the dense correspondence can be reconstructed from estimated $\mathbf{C}^{\ast}$ by conducting nearest neighbor search between the rows of $\Phi_x \mathbf{C}_{yx}$ and that of $\Phi_y$, with possible post-processing~\cite{melzi2019zoomout, ezuz2017deblurring, ren2018continuous}.  

\subsection{Efficient Optimal Transport}
\label{subsec:efficient-OT}
\textbf{Wasserstein distance.} For $p\geq 1$, given two probability measures $\mu \in \mathcal{P}_p(\mathbb{R}^d)$ and $\nu \in \mathcal{P}_p(\mathbb{R}^d)$, the Wasserstein distance~\cite{villani2003topics} between $\mu$ and $\nu$ is : 
\begin{align}
\label{eq:W}
    \text{W}_p^p(\mu,\nu)  = \inf_{\pi \in \Pi(\mu,\nu)} \int_{\mathbb{R}^d \times \mathbb{R}^d} \| x - y\|_p^{p} d \pi(x,y),
\end{align}
where $\Pi(\mu,\nu)$ are the set of all couplings between $\mu$ and $\nu$ i.e., joint probability measures that have marginals as $\mu$ and $\nu$ respectively. The Wasserstein distance is the optimal transportation cost between $\mu$ and $\nu$ since it is computed with the optimal coupling. As mentioned in the introduction section, the downside of Wasserstein distance is a high computational complexity in the discrete case i.e., $\mathcal{O}(m^3\log m)$ in time and $\mathcal{O}(m^2)$ in space for $m$ is the number of supports. To reduce the time complexity, entropic regularized optimal transport~\cite{cuturi2013sinkhorn} is introduced. 

\vspace{0.5 em}
\noindent
\textbf{Sinkhorn divergence.} For $p\geq 1$, given two probability measures $\mu \in \mathcal{P}_p(\mathbb{R}^d)$ and $\nu \in \mathcal{P}_p(\mathbb{R}^d)$, the Sinkhorn-p divergence~\cite{cuturi2013sinkhorn} between $\mu$ and $\nu$ is : 

\begin{align}
    \label{eq:sinkhorn}
    \text{S}_{\epsilon,p}^p(\mu,\nu) = \inf_{\pi \in \Pi_\epsilon(\mu,\nu)} \int_{\mathbb{R}^d \times \mathbb{R}^d} c d \pi(x,y) + \epsilon H(\pi), 
\end{align}

where $\Pi_\epsilon(\mu,\nu) =\{\pi \in \Pi(\mu,\nu)|\text{KL}(\pi,\mu \otimes \nu) \leq \epsilon\}$ with KL denotes the Kullback Leibler divergence. The cost $c: \mathbb{R}^d \times \mathbb{R}^d \mapsto \mathbb{R}$ is defined as $c_p(x,y) = \|x - y \|_p^p$ on $\mathbb{R}^d \times \mathbb{R}^d$. The entropy term $H(\pi)$ allows us to solve for the correspondence $\pi$ via Sinkhorn-Knopp algorithm with $\mathcal{O}(m^2)$ in time complexity.

\vspace{0.5 em}
\noindent
\textbf{Sliced Wasserstein distance.} The sliced Wasserstein (SW) distance~\cite{bonneel2015sliced} between two probability measures $\mu \in \mathcal{P}_p(\mathbb{R}^d)$ and $\nu\in \mathcal{P}_p(\mathbb{R}^d)$ is given by:
\begin{align}
\label{eq:SW}
    \text{SW}_p^p(\mu,\nu)  = \mathbb{E}_{ \theta \sim \mathcal{U}(\mathbb{S}^{d-1})} [\text{W}_p^p (\theta \sharp \mu,\theta \sharp \nu)],
\end{align}
where $\theta \sharp \nu$ denotes the push-forward measure of $\mu$ via function $f(x) = \theta^\top x$, and  the one-dimensional Wasserstein distance appears in a closed form which is $\text{W}_p^p(\theta \sharp \mu,\theta \sharp \nu) =
     \int_0^1 |F_{\theta \sharp \mu}^{-1}(z) - F_{\theta \sharp \nu}^{-1}(z)|^{p} dz $. Here, $F_{\theta \sharp \mu}$ and $F_{\theta \sharp \nu}$  are  the cumulative
distribution function (CDF) of $\theta \sharp \mu$ and $\theta \sharp \nu$ respectively. When $\mu$ and $\nu$ have at most $n$ supports, the computation of the SW is only $\mathcal{O}(n\log n)$ in time and $\mathcal{O}(n)$ in space. The SW often is computed by using $L$ Monte Carlo samples $\theta_1,\ldots,\theta_L$ from the unit sphere:
\begin{align}
\label{eq:MCSW}
    \widehat{\text{SW}}_{p}^{p}(\mu,\nu;L) = \frac{1}{L}\sum_{l=1}^L\text{W}_p^p (\theta_l \sharp \mu,\theta_l \sharp \nu).
\end{align}
\textbf{Energy-based Sliced Wasserstein distance.} Energy-based sliced Wasserstein (EBSW) is a more discriminative variant of the SW proposed in~\cite{nguyen2023energy}. The definition of the EBSW is given as:
\begin{align}
        \text{EBSW}_p^p(\mu,\nu;f) = \mathbb{E}_{\theta \sim \sigma_{\mu,\nu}(\theta;f,p)}\left[ \text{W}_p^p (\theta\sharp \mu,\theta \sharp \nu)\right],
    \end{align}
where $f$ is the energy function e.g., $f(x)=e^x$, and $\sigma_{\mu,\nu}(\theta;f,p) \propto f(\text{W}_p^p(\theta \sharp \mu,\theta \sharp \nu)) \in \mathcal{P}(\mathbb{S}^{d-1})$ is the energy-based slicing distribution. The EBSW can be computed based on importance sampling with $L$ samples from proposal distribution $\sigma_0(\theta)$, e.g., $\mathcal{U}(\mathbb{S}^{d-1})$. For $\theta_1,\ldots,\theta_L \overset{i.i.d}{\sim} \sigma_0(\theta)$, we have:
\begin{align}
\label{eq:emp_ISEBSW}
   \widehat{\text{IS-EBSW}}_p^p(\mu,\nu;f,L) & \nonumber \\
   & \hspace{- 4 em} =  \sum_{l=1}^L   \text{W}_p^p (\theta_l\sharp \mu,\theta_l \sharp \nu)\hat{w}_{\mu,\nu,\sigma_0,f ,p} (\theta_l) ,
\end{align}
for $w_{\mu,\nu,\sigma_0,f,p } (\theta) = \frac{f(\text{W}^p_p(\theta \sharp \mu,\theta \sharp \nu) )}{\sigma_0(\theta)}$ is the importance weighted function and $\hat{w}_{\mu,\nu,\sigma_0,f ,p} (\theta_l) = \frac{w_{\mu,\nu,\sigma_0,f,p } (\theta_l)}{\sum_{l'=1}^L w_{\mu,\nu,\sigma_0,f,p } (\theta_{l'})}$ is the normalized importance weights.

\section{Learning Shape Correspondence with Efficient Optimal Transport}
\label{sec:proposed-method}

In this section, we provide in-depth details of our proposed non-rigid shape matching framework. The whole framework is described in Fig.~\ref{fig:framework}. Our pipeline starts by extracting features from the feature extractor as described in Sec.~\ref{subsec:feature-extractor}. Then we describe functional map in Sec.~\ref{subsec:FM-method}. Thirdly, we illustrate how efficient OT in Sec.~\ref{subsec:feature-extrinsic-OT} is applied to our framework and propose two novel loss functions for learning precise shape mapping. Thirdly, we summarize our unsupervised losses in Sec.~\ref{subsec:loss-functions}. Finally, we propose an adaptive refinement process in Sec.~\ref{subsec:adaptive-refinement}.

\subsection{Feature extractor}
\label{subsec:feature-extractor}
Our architecture is designed in the form of a Siamese network. Specifically, we utilize the same feature extractor with shared learning parameters to extract features from a pair of input shapes. We employ DiffusionNet~\cite{sharp2020diffusionnet} as our feature extractor since DiffusionNet is agnostic to discretization and resolution of the meshes, thereby ensuring robust shape correspondence. Consequently, from the pair of inputs, we extract two sets of features, denoted by $\mathcal{F}_x \in \mathbb{R}^{n_x \times d}$ and $\mathcal{F}_y \in \mathbb{R}^{n_y \times d}$ via DiffusionNet. 

\subsection{Functional map module}
\label{subsec:FM-method}
As discussed in~\ref{subsec:FM}, we aim to employ deep functional map as a proxy to learn an intrinsic feature shape matching. Specifically, we employ regularized functional map~\cite{ren2019structured}, to compute optimal functional map $\mathbf{C}^{\ast}$ as mentioned in Sec.~\ref{subsec:FM}. During training, the network aims to minimize the structural regularization of functional map:
\begin{equation}
    \label{eq:l_fmap}
    \mathcal{L}_{fmap} = \alpha_1 \mathcal{L}_{bij} + \alpha_2 \mathcal{L}_{othor},
\end{equation}
where $\mathcal{L}_{bij} = \|C_{xy}C_{yx} - I \|^2 + \|C_{yx}C_{xy} - I \|^2$ promotes identity mapping and $\mathcal{L}_{othor} = \|C_{xy}^TC_{yx} - I \|^2 + \|C_{yx}^TC_{xy} - I \|^2$ imposes locally area-preserving~\cite{ren2019structured}. 

\subsection{Feature extrinsic alignment via efficient optimal transport}
\label{subsec:feature-extrinsic-OT}
We aim to integrate efficient OT into deep functional map to promote precise mesh feature alignment. Thanks to the fast computation and the closed-form solution of sliced Wasserstein (SW) distance, we derive a novel loss function based on SW distance. 

\vspace{0.5 em}
\noindent
\textbf{Soft feature similarity.} Firstly, from a pair of features $\mathcal{F}_x, \mathcal{F}_y$ extracted from shapes $\mathcal{X}, \mathcal{Y}$, respectively, we estimate a \textit{soft feature similarity matrix} $\hat{\Pi}_{xy} \in \mathbb{R}^{n_x \times n_y}$ such that:
\begin{equation}
    \label{eq:soft-similarity}
    \hat{\Pi}_{xy}^{i, j} = \frac{\text{exp}((\mathcal{F}^i_x \cdot \mathcal{F}^j_y)/ \tau)}{\sum_{k=1}^{n_y} \text{exp}((\mathcal{F}^i_x \cdot \mathcal{F}^k_y)/ \tau))},
\end{equation}
where $\tau$ is scaling factor, and $\mathcal{F}^i_x, \mathcal{F}^j_y \in \mathbb{R}^d$ represent $d$-dimensional features of point $i^{th}$ in shape $\mathcal{X}$ and $j^{th}$ in shape $\mathcal{Y}$, respectively. Similarly, the $\hat{\Pi}_{yx}$ is constructed in the same fashion as in Eq.~\ref{eq:soft-similarity}. 


\vspace{0.5 em}
\noindent
\textbf{Feature alignment via OT.} Finding precise point-to-point mapping based on feature similarity requires solving linear assignment problem in $\mathbb{R}^d$, which is expensive to integrate into a learning-based framework. Therefore, in this work, we relax the constraints to cast the feature-matching problem as a probability distribution matching problem. In other words, we represent the extracted features $\mathcal{F}_x, \mathcal{F}_y$ as probability distributions defined over $\mathbb{R}^d$. After that, we attempt to learn mappings that minimize the ``distance" between the two distributions, i.e. probability measures. The OT cost~\cite{villani2021topics} is a naturally fitted discrepancy between probability measures, thereby being employed in our framework. 

\vspace{0.5 em}
\noindent
\textbf{SW distance as an efficient OT.} Thanks to the fast computation and its closed-form solution of SW distance, we derive a novel loss function that jointly learns the mapping and minimizes the discrepancy between two feature probability measures as follows:
\begin{equation}
\begin{aligned}
    \label{eq:sw-align}
    \mathcal{L}_{biSW}  = (\mathbb{E}_{ \theta \sim \mathcal{U}(\mathbb{S}^{d-1})} [\text{W}_p^p (\theta \sharp \mathcal{F}_x,\theta \sharp \hat{\mathcal{F}}_y) & \\ & \hspace{- 6 em} + \text{W}_p^p (\theta \sharp \mathcal{F}_y,\theta \sharp \hat{\mathcal{F}}_x)])^{\frac{1}{p}},
\end{aligned}
\end{equation}
where $\hat{\mathcal{F}}_x = \hat{\Pi}_{yx} \mathcal{F}_x$ and $\hat{\mathcal{F}}_y = \hat{\Pi}_{xy} \mathcal{F}_y$. The loss $\mathcal{L}_{biSW}$ minimizes the discrepancy between the feature probability measures in one shape and the softly permuted feature sets of its counterpart in a bidirectional manner. The loss converges toward zero when the soft feature similarity $\hat{\Pi}$ approaches a (partial) permutation matrix, indicating that the point-wise corresponding features are closely aligned. Moreover, the loss encourages the cycle consistency of the mapping.  It is worth noting that our loss diverges from contrastive losses explored in prior works~\cite{li2022learning, xie2020pointcontrast, cao2023self}. Where the contrastive loss only considers whether individual point correspondences are correct or not, our proposed loss introduces a more general and flexible matching by conceptualizing the point features as probability measures and employing OT cost as a metric of evaluation. 

\vspace{0.5 em}
\noindent
\textbf{Bidirectional EBSW. } It is worth noting that the proposed loss $\mathcal{L}_{biSW}$ in Eq.~\ref{eq:sw-align} employs the projecting directions sampled from uniform distribution over unit-hypersphere as the shared slicing distributions. Despite being easy to sample, the uniform distribution is not able to differentiate between informative and non-informative projecting features. Therefore, inspired by~\cite{nguyen2023energy}, we propose a bidirectional energy-based SW loss defined in the importance sampling form as:
\begin{equation}
\begin{aligned}
    \label{eq:ISEBSW-align}
    \mathcal{L}_{biEBSW} = \left( \frac{\mathbb{E}_{\theta \sim \sigma_0(\theta)} [(\text{W}_{\theta, \mathcal{X}}+ \text{W}_{\theta, \mathcal{Y}}) w(\theta) ]}
                            {\mathbb{E}_{\theta \sim \sigma_0(\theta)} [w(\theta)]} \right)^{\frac{1}{p}},
\end{aligned}
\end{equation}
where we denote $\text{W}_{\theta, \mathcal{X}} \coloneqq \text{W}_p^p (\theta \sharp \mathcal{F}_x,\theta \sharp \hat{\mathcal{F}}_y), \text{W}_{\theta, \mathcal{Y}} \coloneqq \text{W}_p^p (\theta \sharp \mathcal{F}_y,\theta \sharp \hat{\mathcal{F}}_x)$, and $w(\theta) \coloneqq \frac{\text{exp}(\text{W}_{\theta, \mathcal{X}} + \text{W}_{\theta, \mathcal{Y}})}{\sigma_0(\theta)}$. The loss $\mathcal{L}_{biEBSW}$ shares the same properties for shape correspondence as the vanilla SW loss in Eq.~\ref{eq:sw-align}. However, it imposes a more expressive mechanism for selecting projection directions in the computation of the SW distance. Moreover, the vanilla SW loss can be seen as a summation of two SW distances since the slicing distribution is fixed as uniform. In contrast, the bidirectional EBSW loss has the slicing distribution shared and affected by both one-dimensional Wasserstein distances. Hence, the bidirectional EBSW is considerably different from the original EBSW  in~\cite{nguyen2023energy}.

We provide detailed computation and discussion of $\mathcal{L}_{biSW}$ and $\mathcal{L}_{biEBSW}$ at Sup.~\ref{sec:additional-materials}.

\subsection{Loss functions}
\label{subsec:loss-functions}
\textbf{Proper functional maps.} We employ the notion of proper functional map introduced by~\cite{ren2021discrete}: \textit{The functional map $C_{xy}$  is deemed ``proper" if there exists a (partial) permutation matrix $\Pi_{yx}$ so that $C_{xy} = \Phi_y^{\dagger} \Pi_{yx} \Phi_x$.} Drawing on this concept, we introduce a loss term that not only promotes the ``properness" of the functional map but also concurrently regularizes the (OT) cost, namely:

\begin{equation}
    \label{eq:proper-loss}
    \mathcal{L}_{proper} = \| C_{xy} - \Phi_y^{\dagger} \hat{\Pi}_{yx} \Phi_x \|^2
\end{equation}
It is worth noting that while our $\mathcal{L}_{proper}$ might bear resemblance to the coupling loss in~\cite{cao2023unsupervised}, the proposed loss diverges by using soft feature similarity $\hat{\Pi}_{yx}$ jointly optimized with the feature extrinsic alignment through OT as discussed in Sec.~\ref{subsec:feature-extrinsic-OT}. Therefore, it serves as a strong regularization for imposing structural smoothness of functional map and promoting precise mapping via OT. 

\vspace{0.5 em}
\noindent
\textbf{Total loss.} Our framework is trained end-to-end without annotation by minimizing the following unsupervised losses:
\begin{equation}
    \label{eq:total-losses}
    \mathcal{L}_{total} = \lambda_{1} \mathcal{L}_{fmap} + \lambda_{2} \mathcal{L}_{OT} + \lambda_{3} \mathcal{L}_{proper},
\end{equation}
where $\lambda_i$ is the weight for each loss, and $\mathcal{L}_{OT}$ could be either $\mathcal{L}_{biSW}$ or $\mathcal{L}_{biEBSW}$.

\subsection{Adaptive refinement via entropic optimal transport}
\label{subsec:adaptive-refinement}
\textbf{Adaptive refinement. }To provide a more precise correspondence, we propose an adaptive refinement module designed to incrementally improve the final match for each individual shape pairing. Specifically, we estimate the pseudo soft correspondence $\tilde{\Pi}$ via entropic regularized optimal transport~\cite{cuturi2013sinkhorn} as mentioned in Eq.~\ref{eq:sinkhorn} is defined as:
\begin{equation}
    \label{eq:sinkhorn-adaptive}
    \tilde{\Pi}_{xy} = \mathcal{Q}^ {\mathcal{X}} ( \mathcal{Q}^{\mathcal{Y}}\cdots (\mathcal{Q}^{\mathcal{X}} (p_{\epsilon}) ) ),
\end{equation}
where $\mathcal{Q}(\cdot)$ is the projection operator of a given probability density $p: \mathbb{R}^d \times \mathbb{R}^d \rightarrow \mathbb{R}$ defined as: $p_{\epsilon} (x,y) \propto \exp(- \frac{1}{\epsilon} c_2(x,y))$. Thanks to the differentiable property of the Sinkhorn algorithm, we can refine each individual pair by minimizing the $\mathcal{L}_{total}$ to update the features accordingly. In contrast to the axiomatic method~\cite{melzi2019zoomout} that often requires alternately updating the functional map and pointwise map, our method offers a differentiable process that facilitates simultaneous updates. Furthermore, it is noteworthy that our approach is orthogonal to~\cite{eisenberger2020deep} since we only employ entropic OT for refinement once during the inference, thereby reducing the computation and memory cost of the Sinkhorn algorithm. We provide detailed algorithms of adaptive refinement at Sup.~\ref{sec:additional-materials}.

\vspace{0.5 em}
\noindent
\textbf{Inference. } During inference, our final mapping is obtained by nearest neighbor search on features extracted from the feature extractor module.
\begin{table*}[!ht]
\centering

\setlength{\tabcolsep}{6pt}
    \centering
    \caption{\textbf{Quantitative results on near-isometric shape matching.} The color denotes the \colorbox{tabfirst}{best} and \colorbox{tabsecond}{second}-best result. Our method outperforms various methods including axiomatic, supervised and unsupervised methods in most settings.}
    
    \resizebox{\linewidth}{!}{%
    \label{tab:near-isometric}
        \begin{tabular}{@{}lccccccccc@{}}
        \toprule
        \multicolumn{1}{l}{Method}  & \multicolumn{3}{c}{\textbf{FAUST}}   & \multicolumn{3}{c}{\textbf{SCAPE}}  & \multicolumn{3}{c}{\textbf{FAUST + SCAPE}} \\ \cmidrule(lr){2-4} \cmidrule(lr){5-7} \cmidrule(lr){8-10}
        \multicolumn{1}{l}{} & \multicolumn{1}{c}{\textbf{FAUST}} & \multicolumn{1}{c}{\textbf{SCAPE}} & \multicolumn{1}{c}{\textbf{SHREC'19}} & \multicolumn{1}{c}{\textbf{FAUST}} & \multicolumn{1}{c}{\textbf{SCAPE}} & \multicolumn{1}{c}{\textbf{SHREC'19}} & \multicolumn{1}{c}{\textbf{FAUST}} & \multicolumn{1}{c}{\textbf{SCAPE}} & \multicolumn{1}{c}{\textbf{SHREC'19}}
        \\ \midrule
        \multicolumn{10}{c}{\underline{Axiomatic}} \\
        \multicolumn{1}{l}{ZoomOut~\cite{melzi2019zoomout}} & \multicolumn{1}{c}{6.1}  & \multicolumn{1}{c}{\textbackslash{}} & \multicolumn{1}{c}{\textbackslash{}}  & \multicolumn{1}{c}{\textbackslash{}} & \multicolumn{1}{c}{7.5} & \multicolumn{1}{c}{\textbackslash{}} & \multicolumn{1}{c}{\textbackslash{}}    & \multicolumn{1}{c}{\textbackslash{}} & \multicolumn{1}{c}{\textbackslash{}}\\
        
        \multicolumn{1}{l}{BCICP~\cite{ren2018continuous}}  & \multicolumn{1}{c}{6.1}  & \multicolumn{1}{c}{\textbackslash{}} & \multicolumn{1}{c}{\textbackslash{}}  & \multicolumn{1}{c}{\textbackslash{}} & \multicolumn{1}{c}{11.0} & \multicolumn{1}{c}{\textbackslash{}} & \multicolumn{1}{c}{\textbackslash{}}    & \multicolumn{1}{c}{\textbackslash{}} & \multicolumn{1}{c}{\textbackslash{}}\\
        
        \multicolumn{1}{l}{Smooth Shells~\cite{eisenberger2020smooth}} & \multicolumn{1}{c}{2.5}  & \multicolumn{1}{c}{\textbackslash{}} & \multicolumn{1}{c}{\textbackslash{}}  & \multicolumn{1}{c}{\textbackslash{}} & \multicolumn{1}{c}{4.7} & \multicolumn{1}{c}{\textbackslash{}} & \multicolumn{1}{c}{\textbackslash{}}    & \multicolumn{1}{c}{\textbackslash{}} & \multicolumn{1}{c}{\textbackslash{}}\\ 

        \midrule
        \multicolumn{10}{c}{\underline{Supervised}} \\ 
        \multicolumn{1}{l}{FMNet~\cite{litany2017deep}} & \multicolumn{1}{c}{11.0}  & \multicolumn{1}{c}{30.0} & \multicolumn{1}{c}{\textbackslash{}}  & \multicolumn{1}{c}{33.0} & \multicolumn{1}{c}{17.0} & \multicolumn{1}{c}{\textbackslash{}} & \multicolumn{1}{c}{\textbackslash{}}    & \multicolumn{1}{c}{\textbackslash{}} & \multicolumn{1}{c}{\textbackslash{}} \\

        \multicolumn{1}{l}{GeomFMaps~\cite{donati2020deep}}& \multicolumn{1}{c}{2.6}  & \multicolumn{1}{c}{3.3} & \multicolumn{1}{c}{9.9}  & \multicolumn{1}{c}{3.0} & \multicolumn{1}{c}{3.0} & \multicolumn{1}{c}{12.2} & \multicolumn{1}{c}{2.6}    & \multicolumn{1}{c}{3.0} & \multicolumn{1}{c}{7.9}\\
        
        \multicolumn{1}{l}{TransMatch~\cite{trappolini2021shape}}& \multicolumn{1}{c}{1.8}  & \multicolumn{1}{c}{32.8} & \multicolumn{1}{c}{19.0}  & \multicolumn{1}{c}{18.5} & \multicolumn{1}{c}{16.0} & \multicolumn{1}{c}{39.5} & \multicolumn{1}{c}{\cellcolor{tabsecond}1.7} & \multicolumn{1}{c}{13.5} & \multicolumn{1}{c}{12.9}\\
        
        \midrule
        \multicolumn{10}{c}{\underline{Unsupervised}} \\ 
        \multicolumn{1}{l}{SURFMNet~\cite{roufosse2019unsupervised}} & \multicolumn{1}{c}{15.0}  & \multicolumn{1}{c}{32.0} & \multicolumn{1}{c}{\textbackslash{}}  & \multicolumn{1}{c}{32.0} & \multicolumn{1}{c}{12.0} & \multicolumn{1}{c}{\textbackslash{}} & \multicolumn{1}{c}{33.0}    & \multicolumn{1}{c}{29.0} & \multicolumn{1}{c}{\textbackslash{}} \\

        \multicolumn{1}{l}{Deep Shells~\cite{eisenberger2020deep}} & \multicolumn{1}{c}{1.7}  & \multicolumn{1}{c}{5.4} & \multicolumn{1}{c}{27.4}  & \multicolumn{1}{c}{2.7} & \multicolumn{1}{c}{2.5} & \multicolumn{1}{c}{23.4} & \multicolumn{1}{c}{\cellcolor{tabfirst}1.6}   & \multicolumn{1}{c}{2.4} & \multicolumn{1}{c}{21.1} \\

        \multicolumn{1}{l}{AFMap~\cite{li2022learning}}  & \multicolumn{1}{c}{1.9}  & \multicolumn{1}{c}{\cellcolor{tabfirst}2.6} & \multicolumn{1}{c}{\cellcolor{tabsecond}6.4}  & \multicolumn{1}{c}{2.2} & \multicolumn{1}{c}{2.2} & \multicolumn{1}{c}{9.9} & \multicolumn{1}{c}{1.9}    & \multicolumn{1}{c}{\cellcolor{tabsecond}2.3} & \multicolumn{1}{c}{5.8}\\

        \multicolumn{1}{l}{SSLMSM~\cite{cao2023self}}  & \multicolumn{1}{c}{2.0}  & \multicolumn{1}{c}{7.0} & \multicolumn{1}{c}{9.1}  & \multicolumn{1}{c}{2.7} & \multicolumn{1}{c}{3.1} & \multicolumn{1}{c}{8.4} & \multicolumn{1}{c}{1.9}    & \multicolumn{1}{c}{4.3} & \multicolumn{1}{c}{6.2} \\
        
        \multicolumn{1}{l}{UDMSM~\cite{cao2022unsupervised}} & \multicolumn{1}{c}{\cellcolor{tabfirst}1.5}  & \multicolumn{1}{c}{7.5} & \multicolumn{1}{c}{20.1}  & \multicolumn{1}{c}{3.2} & \multicolumn{1}{c}{2.0} & \multicolumn{1}{c}{28.3} & \multicolumn{1}{c}{\cellcolor{tabsecond}1.7}    & \multicolumn{1}{c}{7.6} & \multicolumn{1}{c}{28.7} \\

        \multicolumn{1}{l}{ULRSSM~\cite{cao2023unsupervised}} & \multicolumn{1}{c}{1.6}  & \multicolumn{1}{c}{3.6} & \multicolumn{1}{c}{7.2}  & \multicolumn{1}{c}{\cellcolor{tabsecond}1.9} & \multicolumn{1}{c}{\cellcolor{tabsecond}\cellcolor{tabsecond}1.9} & \multicolumn{1}{c}{\cellcolor{tabsecond}7.6} & \multicolumn{1}{c}{\cellcolor{tabsecond}1.7}    & \multicolumn{1}{c}{3.2} & \multicolumn{1}{c}{\cellcolor{tabfirst}4.6} \\

        \midrule
        \multicolumn{1}{l}{Ours}  & \multicolumn{1}{c}{\cellcolor{tabfirst}1.5}  & \multicolumn{1}{c}{\cellcolor{tabsecond}3.4} & \multicolumn{1}{c}{\cellcolor{tabfirst}5.5}  & \multicolumn{1}{c}{\cellcolor{tabfirst}1.6} & \multicolumn{1}{c}{\cellcolor{tabfirst}1.8} & \multicolumn{1}{c}{\cellcolor{tabfirst}7.0} & \multicolumn{1}{c}{\cellcolor{tabfirst}1.6}    & \multicolumn{1}{c}{\cellcolor{tabfirst}2.2} & \multicolumn{1}{c}{\cellcolor{tabsecond}4.7} \\\hline
        \end{tabular} 
    }
\end{table*}

\section{Experimental results}
\label{sec:experimental-results}

\textbf{Datasets. }We conduct a series of experiments across diverse shape-matching datasets and their application on a downstream task. Specifically, we perform experiment on human shape matching with near-isometric dataset such as FAUST~\cite{bogo2014faust} and SCAPE~\cite{anguelov2005scape} as well as non-isometric dataset SHREC'19~\cite{melzi2019shrec}. Furthermore, our study extends to two non-isometric animal datasets: SMAL~\cite{zuffi20173d} and the more recent DeformingThings4D~\cite{li20214dcomplete, magnet2022smooth}. Finally, we conclude our experiments by performing segmentation transfer on 3D semantic segmentation dataset introduced in~\cite{abdelreheem2023satr}.


\vspace{0.5 em}
\noindent
\textbf{Baselines. }We conduct extensive comparisons with a wide range of non-rigid shape matching methods: (1) Axiomatic methods including ZoomOut~\cite{melzi2019zoomout}, BCICP~\cite{ren2018continuous}, Smooth Shells~\cite{eisenberger2020smooth}; (2) Supervised methods including FMNet~\cite{litany2017deep}, GeomFMaps~\cite{donati2020deep}, TransMatch~\cite{trappolini2021shape}; (3) Unsupervised methods including SURFMNet~\cite{roufosse2019unsupervised}, Deep Shells~\cite{eisenberger2020deep}, AFMap~\cite{li2022learning}, SSLMSM~\cite{cao2023self}, UDMSM~\cite{cao2022unsupervised}, ULRSSM~\cite{cao2023unsupervised}. While there are numerous non-rigid shape-matching methods in the literature, we decided to choose the most recent and relevant to our works for comparison.  

\vspace{0.5 em}
\noindent
\textbf{Metrics. }Regarding shape matching metric, similar to all of our competing methods, we employ mean geodesic errors ($\times 100$)~\cite{kim2011blended}. For segmentation transfer, we use semantic segmentation mIOU as in~\cite{kim2020large}.

\subsection{Near-isometric Shape Matching}
\label{subsec:near-iso}

\textbf{Datasets. }We employ a more challenging remeshed version of FAUST~\cite{bogo2014faust} and SCAPE~\cite{anguelov2005scape}, as proposed in~\cite{donati2020deep, ren2018continuous}. The remeshed FAUST dataset includes $100$ shapes, representing $10$ individuals in $10$ different poses, with the evaluation focusing on the final $20$ shapes. Similarly, the remeshed SCAPE dataset comprises $71$ poses of a single individual, where again, the last $20$ shapes are used for evaluation purposes. Additionally, the SHREC’$19$ dataset presents a more complex challenge due to its significant variations in mesh connectivity, encompassing $44$ shapes and $430$ pairs for evaluation.

\begin{figure}
\centering
\includegraphics[width=0.47\textwidth]{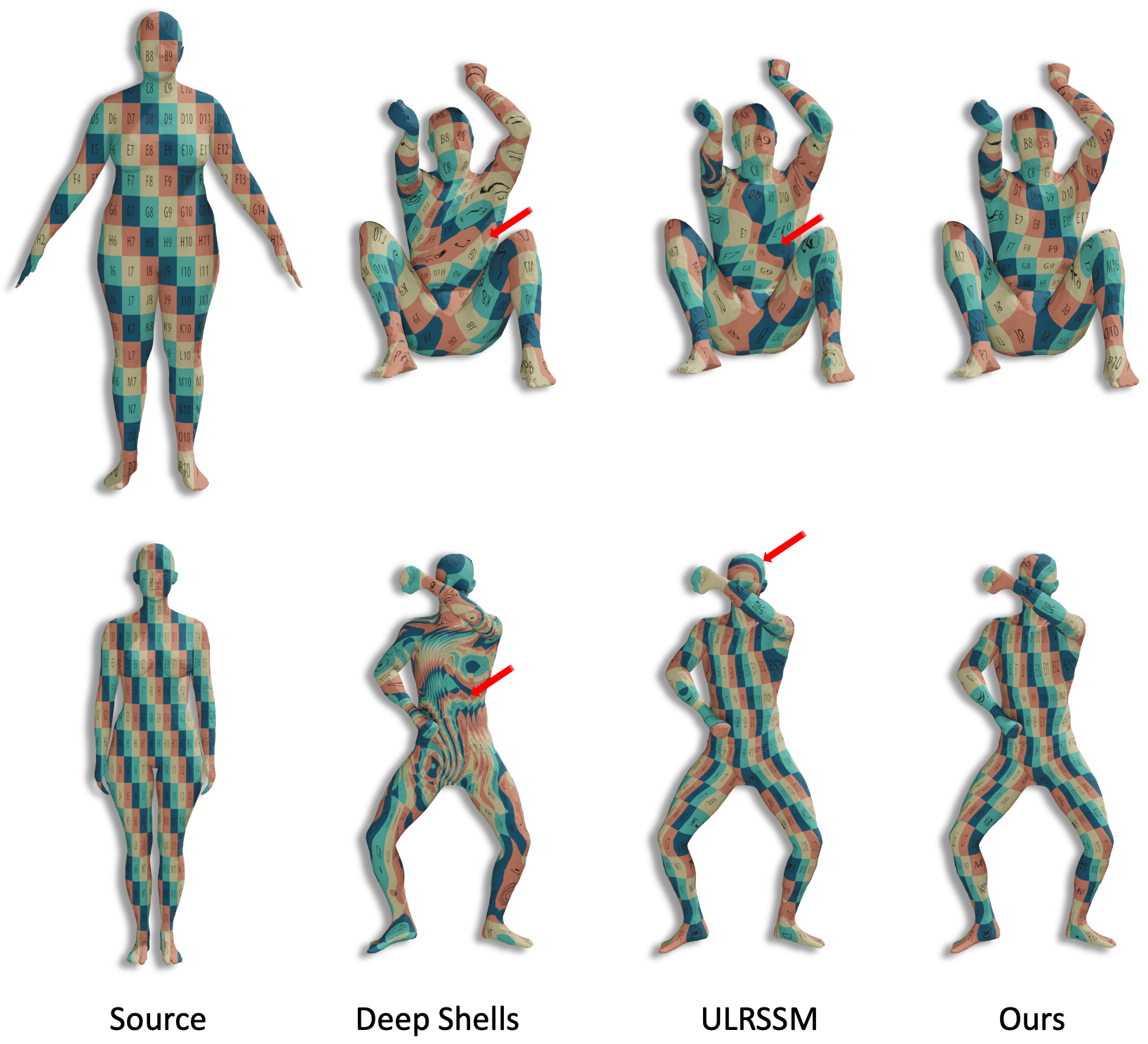}
\caption{\textbf{Qualitative results} of different methods evaluated on SHREC'19 datasets. Correspondence is visualized by texture transfer. The red arrow indicates poor mappings.}
\label{fig:shrec-qualitative}
\end{figure}

\vspace{0.5 em}
\noindent
\textbf{Results. }We conduct experiments on FAUST, SCAPE, and the combination of both datasets. Quantitative results in Tab.~\ref{tab:near-isometric} show that supervised methods tend to overfit the trained dataset. On the other hand, unsupervised methods typically can achieve a better generalization on new datasets. Compared to Deep Shells, an OT-based method, we outperform in most settings as shown in Tab.~\ref{tab:near-isometric} and Fig.~\ref{fig:shrec-qualitative}. Compared to state-of-the-art ULRSSM, our method indicates a slightly better mapping demonstrated in Fig.~\ref{fig:shrec-qualitative}.

\subsection{Non-isometric Shape Matching}
\label{subsec:non-iso}
\begin{table}[h]
    \setlength{\tabcolsep}{9pt}
    \small
    \centering
    \caption{\textbf{Quantitative results for non-isometric matching on SMAL and DT4D-H.} Our method surpass state-of-the-art methods on challenging non-isometric dataset such as SMAL and DT4D-H.}
    \label{tab:non-isometric}
    \resizebox{\columnwidth}{!}{
    \begin{tabular}{@{}lcc@{}}
    \toprule
    \multicolumn{1}{c}{Method}      & \multicolumn{1}{c}{\textbf{SMAL}} & \multicolumn{1}{c}{\textbf{DT4D-H}}

    \\\midrule
    \multicolumn{1}{l}{Deep Shells~\cite{eisenberger2020deep}}  & \multicolumn{1}{c}{29.3} & \multicolumn{1}{c}{31.1}\\
    \multicolumn{1}{l}{GeoFMaps~\cite{donati2020deep}}  & \multicolumn{1}{c}{7.6} & \multicolumn{1}{c}{22.6}\\
    \multicolumn{1}{l}{AFMap~\cite{li2022learning}}  & \multicolumn{1}{c}{5.4} & \multicolumn{1}{c}{11.6}\\
    \multicolumn{1}{l}{ULRSSM~\cite{cao2023unsupervised}}  & \multicolumn{1}{c}{\cellcolor{tabsecond}4.2} & \multicolumn{1}{c}{\cellcolor{tabsecond}4.5} \\
    
    \midrule
    \multicolumn{1}{l}{Ours}  & \multicolumn{1}{c}{\cellcolor{tabfirst}4.0} & \multicolumn{1}{c}{\cellcolor{tabfirst}4.2} \\ \hline
    \end{tabular} 
    }
\end{table}
\begin{figure}[h!]
\centering
\includegraphics[width=0.47\textwidth]{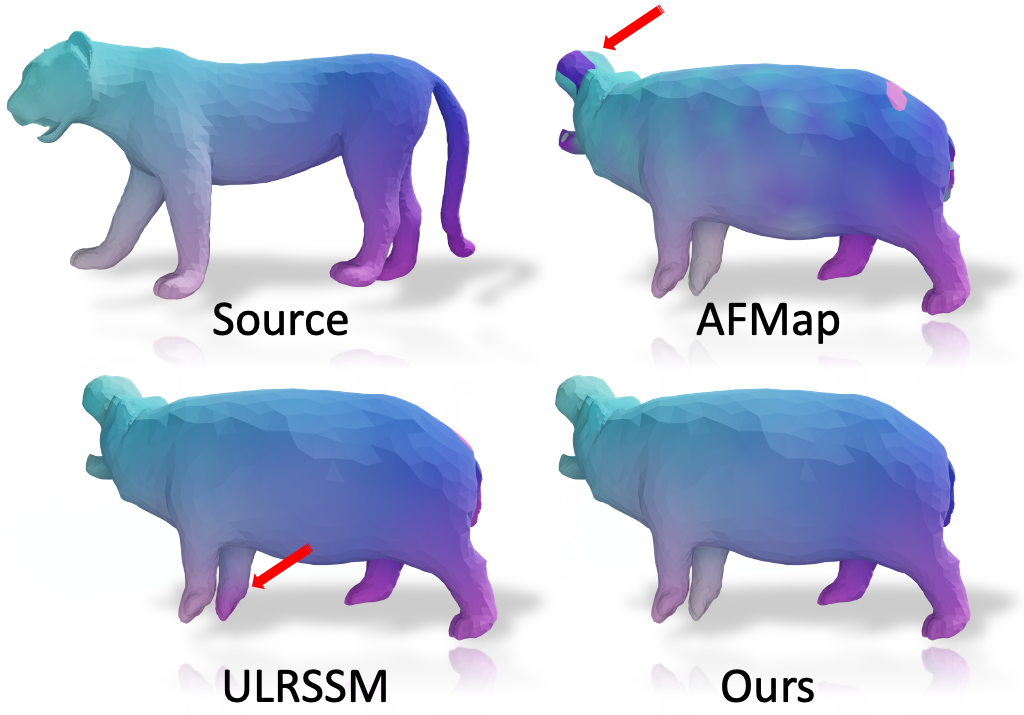}
\caption{\textbf{Qualitative results} of various methods on challenging non-isometric SMAL dataset. Our method demonstrates superior point mapping capabilities compared to previous works. More visualization is provided in Sup.~\ref{sec:additional-visualizations}.}
\label{fig:smal-qualitative}
\vspace{-0.8em}
\end{figure}

\vspace{0.5 em}
\noindent
\textbf{Datasets. }We consider SMAL~\cite{zuffi20173d} and DeformingThings4D~\cite{li20214dcomplete, magnet2022smooth} for evaluating non-isometric shape matching. For the SMAL dataset, we adopt the data split in~\cite{donati2022deep} that uses five species for training and three unseen species for testing, resulting in a $29/20$ split of the dataset. Regarding DeformingThings4D, denoted as DT4D-H, we follow the split also presented in~\cite{donati2022deep} comprising $198$ samples for training and $95$ for testing.

\vspace{0.5 em}
\noindent
\textbf{Results. } To measure the performance on non-isometric datasets, i.e. SMAL and DT4D-H, we compare our method with previous state-of-the-art baselines as shown in Tab.~\ref{tab:non-isometric}. Regarding the DT4D-H dataset, we only perform comparisons on the challenging intra-class scenario. Our proposed method outperforms previous methods in both dataset as shown in Tab.~\ref{tab:non-isometric}. Visualization in Fig.~\ref{fig:smal-qualitative} shows that AFMap often fails to retrieve a non-isometric mapping. In addition, ULRSSM demonstrates better mapping despite some ambiguity. On the other hand, our method obtains a precise and smooth mapping, thus visually better than the two state-of-the-art methods.

\subsection{Segmentation transfer}
\vspace{-0.8em}
\begin{table}[h]
\setlength{\tabcolsep}{14pt}
    \centering
    \small
    \caption{\textbf{Quantitative results for 3D shape segmentation transfer.} Our method is effectively applied to semantic segmentation transfer on 3D shapes, establishing a new benchmark for state-of-the-art performance in this domain.}
    \label{tab:segmentation-transfer}
    \resizebox{\columnwidth}{!}{
    \begin{tabular}{@{}lcc@{}}
    \toprule
    \multicolumn{1}{c}{\textbf{Method}}      & \multicolumn{1}{c}{\textbf{Coarse}} & \multicolumn{1}{c}{\textbf{Fine-grained}}
    \\ \midrule
    
    \multicolumn{1}{l}{AFMaps~\cite{li2022learning}}  & \multicolumn{1}{c}{81.3} & \multicolumn{1}{c}{43.2} \\
    \multicolumn{1}{l}{UDMSM~\cite{cao2022unsupervised}}  & \multicolumn{1}{c}{\cellcolor{tabsecond}85.3} & \multicolumn{1}{c}{45.2} \\
    \multicolumn{1}{l}{ULRSSM~\cite{cao2023unsupervised}}  & \multicolumn{1}{c}{84.2} & \multicolumn{1}{c}{\cellcolor{tabsecond}58.2}\\
    \midrule
    \multicolumn{1}{l}{Ours}  & \multicolumn{1}{c}{\cellcolor{tabfirst}87.8} & \multicolumn{1}{c}{\cellcolor{tabfirst}60.5} \\ \hline
    \end{tabular} 
}
\end{table}
\begin{figure}

\centering
\includegraphics[width=0.47\textwidth]{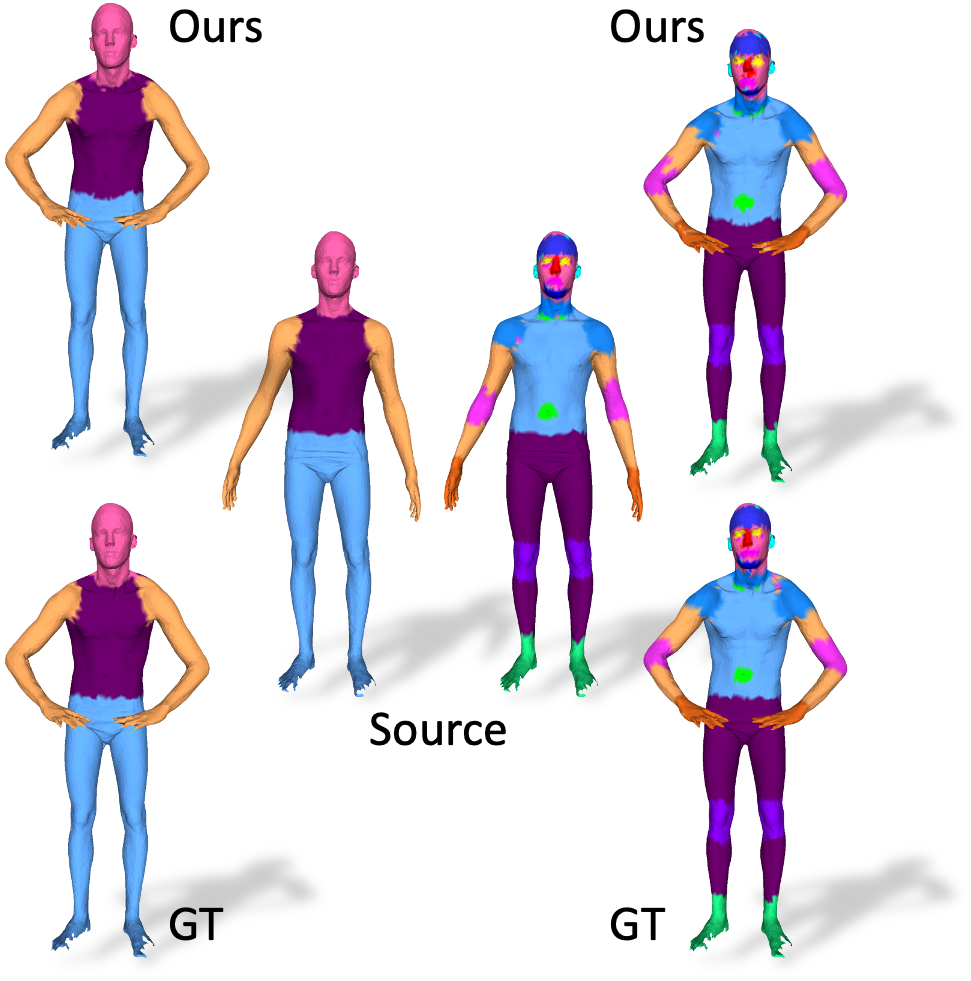}

\caption{\textbf{Qualitative results of segmentation transfer.} Our method exhibits a high-quality segmentation map via computed correspondence. More visualization is provided in Sup.~\ref{sec:additional-visualizations}.}

\label{fig:segmentation-transfer}
\end{figure}

\vspace{0.5 em}
\noindent
\textbf{Datasets. }We illustrate the performance of our proposed method on the task of segmentation transfer on 3D semantic segmentation dataset proposed in~\cite{abdelreheem2023satr}. Specifically, the dataset is derived from FAUST~\cite{bogo2014faust}, which is manually annotated into two types of label: coarse annotations include $4$ classes and fine-grained annotations comprise $17$ categories. After excluding non-connected meshes, we test our method on $79$ meshes by computing correspondence among the collection and then transferring annotation from one single mesh to the others. 

\vspace{0.5 em}
\noindent
\textbf{Results. }To further demonstrate the robustness, we apply our methods on co-segmentation, also known as segmentation transfer task. We train all methods on the remeshed FAUST\_r mentioned in Sec.~\ref{subsec:near-iso}. It is worth noting while the FAUST\_r is remeshed to around $10$K faces, the segmentation dataset in~\cite{abdelreheem2023satr} is remeshed to $20$K triangular faces. Therefore, it showcases the generalization of our method that does not depend on the discretization and resolution of mesh. Tab.~\ref{tab:segmentation-transfer} indicates that our method sets a new state-of-the-art on the segmentation-transfer task on FAUST~\cite{abdelreheem2023satr} dataset in both coarse and fine-grained annotation. Fig.~\ref{fig:segmentation-transfer} shows that our method is very closed to ground truth without the need for training semantic segmentation models.

\section{Ablation study}
\vspace{-0.8em}
\label{sec:ablation-study}
\begin{table}[hbt!]
    \setlength{\tabcolsep}{16pt}
    \centering
    \caption{\textbf{Ablation study on SHREC'19.} In the first setting, we replace $\mathcal{L}_{OT}$ with $\mathcal{L}_{MSE}$ in Eq.~\ref{eq:total-losses}. In the second row, we substitute $\mathcal{L}_{OT}$ with $\mathcal{L}_{uniSW}$. The third row indicates the $\mathcal{L}_{OT}$ being $\mathcal{L}_{biSW}$ as in Eq.~\ref{eq:sw-align}. The fourth row indicates not using adaptive refinement at the end of the training process.}
    \label{tab:ablation}
    \begin{tabular}{@{}ll@{}}
    \toprule
    \multicolumn{1}{l}{\textbf{Ablation Setting}}  & \multicolumn{1}{l}{\textbf{SHREC'19}}
    \\ \midrule
    \multicolumn{1}{l}{w. $\mathcal{L}_{MSE}$}  & 34.3 \\
    \multicolumn{1}{l}{w. $\mathcal{L}_{uniSW}$}  & 4.9 \\
    \multicolumn{1}{l}{w. $\mathcal{L}_{biSW}$}  & 4.8 \\
    \midrule
    \multicolumn{1}{l}{w.o. adaptive refinement}  & 7.2 \\
    \midrule
    \multicolumn{1}{l}{Ours}  & \textbf{4.7} \\ \hline
    \end{tabular} 
\end{table}

\noindent
\textbf{Settings. }We conduct an ablation study to validate our contribution. We train our model on FAUST+SCAPE dataset and evaluate it on SHREC'19 dataset. Firstly, we evaluate the effectiveness of different losses in the feature alignment component. Furthermore, we also investigate the importance of the adaptive refinement module.  

\noindent
\textbf{Results. } Our results are summarized in Tab.~\ref{tab:ablation}. First of all, by comparing the first row with the last row, we conclude that $\mathcal{L}_{MSE}$ can not learn to align features for retrieving point-to-point correspondence. Secondly, we observe that by using bidirectional SW, we can gain a slightly better performance. Finally, the last row indicates that by employing importance sampling energy-based SW, we can even gain better performance.

\section{Conclusion}
\label{sec:conclusion}
In conclusion, we introduce an innovative framework that integrates functional maps with an efficient optimal transport method, notably the sliced Wasserstein distance, to address computational challenges and enhance feature alignment. Our approach significantly outperforms existing methods in non-rigid shape matching across various scenarios, including both near-isometric and non-isometric forms. This advancement, confirmed through successful applications in tasks like segmentation transfer, highlights our method's efficacy and strong generalization potential in shape matching.
{
    \small
    \bibliographystyle{ieeenat_fullname}
    \bibliography{main}
}

\clearpage
\setcounter{page}{1}
\maketitlesupplementary

In this supplementary, we first define some notations that are used in our main paper and supplementary in Sec.~\ref{sec:notations}. We then discuss some limitations of our work and potential future directions to address them in Sec.~\ref{sec:limitation}. In Sec.~\ref{sec:additional-materials}, we provide detailed computation and algorithm to compute the proposed loss functions. Furthermore, we delineate the implementation details and hyperparameters used in our training process in Sec.~\ref{sec:implementation-details}. Finally, we provide additional qualitative results of our proposed approach in Sec.~\ref{sec:additional-visualizations}.

\vspace{-0.8em}
\section{Notations}
\label{sec:notations}
For any $d \geq 2$, we denote $\mathbb{S}^{d-1}:=\{\theta \in \mathbb{R}^{d}\mid  ||\theta||_2^2 =1\}$ and $\mathcal{U}(\mathbb{S}^{d-1})$ as  the unit hyper-sphere and its corresponding  uniform distribution. We denote $\theta \sharp \mu$ as the push-forward measures of $\mu$ through the function $f:\mathbb{R}^{d} \to \mathbb{R}$ that is $f(x) = \theta^\top x$. Furthermore, we denote $\mathcal{P}(\mathcal{X})$ as the set of all probability measures on the set $\mathcal{X}$. For $p\geq 1$, $\mathcal{P}_p(\mathcal{X})$ is the set of all probability measures on the set $\mathcal{X}$ that have finite $p$-moments.

\vspace{-0.8em}
\section{Limitations and discussion}
\label{sec:limitation}

Our work is the first to integrate an efficient optimal transport to functional map framework for shape correspondence, yet it is not without limitations, potentially opening new research directions. First of all, our algorithm is designed for use with clean and complete meshes. An intriguing avenue for future research would be to extend the applicability of our method to more diverse scenarios, such as dealing with partial meshes, noisy point clouds, and other forms of data representation. This expansion would enhance the versatility of our approach in handling a wider range of practical applications. Secondly, our adaptive refinement module, which utilizes an entropic regularized optimal transport for estimating the soft-feature similarity matrix, shows promise in achieving more precise refinement. However, this method is not without its drawbacks, notably a quadratic increase in memory complexity and computational demand. This presents a challenge that future research could address by developing more computationally efficient approximations, thereby making the process more feasible for larger datasets or more resource-constrained environments. Overall, these potential research directions could significantly contribute to the evolution of shape correspondence methodologies. 

\vspace{-0.8em}
\section{Detailed algorithms and discussion}
\label{sec:additional-materials}

\noindent
\textbf{Sliced Wasserstein distance. }The unidirectional sliced Wasserstein distance version of Eq.~\ref{eq:sw-align} is given by: 

\begin{equation}
\begin{aligned}
    \label{eq:uni-sw-align}
    \mathcal{L}_{uniSW}  = (\mathbb{E}_{ \theta \sim \mathcal{U}(\mathbb{S}^{d-1})} \text{W}_p^p (\theta \sharp \mathcal{F}_x,\theta \sharp \hat{\mathcal{F}}_y) )^{\frac{1}{p}},
\end{aligned}
\end{equation}
where $\hat{\mathcal{F}}_y = \hat{\Pi}_{xy} \mathcal{F}_y$. The unidirectional sliced Wasserstein distance given in Eq.~\ref{eq:uni-sw-align} is computed by using $L$ Monte Carlo samples $\theta_1, ..., \theta_L$ from the unit sphere:
\begin{equation}
\begin{aligned}
    \label{eq:uni-sw-MCMC}
    \widehat{\mathcal{L}_{uniSW}}  = \left(\frac{1}{L} \sum_{l=1}^L \text{W}_p^p (\theta_l \sharp \mathcal{F}_x,\theta_l \sharp \hat{\mathcal{F}}_y) \right)^{\frac{1}{p}},
\end{aligned}
\end{equation}
where $\text{W}_p^p(\theta \sharp \mathcal{F}_x,\theta \sharp \hat{\mathcal{F}}_y) =
     \int_0^1 |F_{\theta \sharp \mathcal{F}_x}^{-1}(z) - F_{\theta \sharp \hat{\mathcal{F}}_y}^{-1}(z)|^{p} dz $ denotes the closed form solution one-dimensional Wasserstein distance of two probability measures $\mathcal{F}_x$ and $\hat{\mathcal{F}}_y$. Here, $F_{\theta \sharp \mathcal{F}_x}$ and $F_{\theta \sharp \hat{\mathcal{F}}_y}$  are  the cumulative distribution function (CDF) of $\theta \sharp \mathcal{F}_x$ and $\theta \sharp \hat{\mathcal{F}}_y$ respectively.

Similarly, the bidirectional sliced Wasserstein distance in Eq.~\ref{eq:sw-align} is also estimated by using $L$ Monte Carlo samples $\theta_1, ..., \theta_L$ from the unit sphere:

\begin{equation}
\begin{aligned}
    \label{eq:sw-MCMC}
    \widehat{\mathcal{L}_{biSW}}  = ( \frac{1}{L} \sum_{l=1}^L [\text{W}_p^p (\theta_l \sharp \mathcal{F}_x,\theta_l \sharp \hat{\mathcal{F}}_y) & \\ & \hspace{- 6 em} + \text{W}_p^p (\theta_l \sharp \mathcal{F}_y,\theta_l \sharp \hat{\mathcal{F}}_x)] )^{\frac{1}{p}},
\end{aligned}
\end{equation}
where $\hat{\mathcal{F}}_x = \hat{\Pi}_{yx} \mathcal{F}_x$ and $\hat{\mathcal{F}}_y = \hat{\Pi}_{xy} \mathcal{F}_y$. We provide a pseudo-code for computing the unidirectional and bidirectional sliced Wasserstein distance in Algorithm~\ref{alg:uni-SW} and Algorithm~\ref{alg:SW}, respectively. 

\begin{algorithm*}[!t]
\caption{Computational algorithm of the unidirectional SW distance}
\label{alg:uni-SW}
\begin{algorithmic}

\State \textbf{Input:} Features extracted from feature extractor module $\mathcal{F}_x, \mathcal{F}_y$; $p\geq 1$; soft features similarity $\hat{\Pi}$ from Eq.~\ref{eq:soft-similarity}; and the number of projections $L$.\\
    \hrulefill
  \State Compute $\hat{\mathcal{F}}_y = \hat{\Pi}_{xy} \mathcal{F}_y$
  \For{$l=1$ to $L$}
  \State Sample $\theta_l \sim \mathcal{U}(\mathbb{S}^{d-1})$
  \State Compute $v_l = \text{W}_p^p (\theta_l \sharp \mathcal{F}_x,\theta_l \sharp \hat{\mathcal{F}}_y) $
  \EndFor
  \State Compute $\widehat{\mathcal{L}_{uniSW}} = \left(\frac{1}{L}\sum_{l=1}^L v_l\right)^{\frac{1}{p}}$
 \State \textbf{Return:} $\widehat{\mathcal{L}_{uniSW}}$
\end{algorithmic}
\end{algorithm*}

\begin{algorithm*}[!t]
\caption{Computational algorithm of the bidirectional SW distance}
\label{alg:SW}
\begin{algorithmic}

\State \textbf{Input:} Features extracted from feature extractor module $\mathcal{F}_x, \mathcal{F}_y$; $p\geq 1$; soft features similarity $\hat{\Pi}$ from Eq.~\ref{eq:soft-similarity}; and the number of projections $L$.\\
    \hrulefill
  \State Compute $\hat{\mathcal{F}}_x = \hat{\Pi}_{yx} \mathcal{F}_x$ and $\hat{\mathcal{F}}_y = \hat{\Pi}_{xy} \mathcal{F}_y$
  \For{$l=1$ to $L$}
  \State Sample $\theta_l \sim \mathcal{U}(\mathbb{S}^{d-1})$
  \State Compute $v_l = \text{W}_p^p (\theta_l \sharp \mathcal{F}_x,\theta_l \sharp \hat{\mathcal{F}}_y) + \text{W}_p^p (\theta_l \sharp \mathcal{F}_y,\theta_l \sharp \hat{\mathcal{F}}_x)$
  \EndFor
  \State Compute $\widehat{\mathcal{L}_{biSW}} = \left(\frac{1}{L}\sum_{l=1}^L v_l\right)^{\frac{1}{p}}$
 \State \textbf{Return:} $\widehat{\mathcal{L}_{biSW}}$
\end{algorithmic}
\end{algorithm*}

\begin{algorithm*}[!t]
\caption{Computational algorithm of the unidirectional EBSW distance}
\label{alg:uni-ISEBSW}
\begin{algorithmic}

\State \textbf{Input:} Features extracted from feature extractor module $\mathcal{F}_x, \mathcal{F}_y$; $p\geq 1$; soft features similarity $\hat{\Pi}$ from Eq.~\ref{eq:soft-similarity}; and the number of projections $L$.\\
  \hrulefill
  \State Compute $\hat{\mathcal{F}}_y = \hat{\Pi}_{xy} \mathcal{F}_y$
  \For{$l=1$ to $L$}
  \State Sample $\theta_l \sim \mathcal{U}(\mathbb{S}^{d-1})$
  \State Compute $v_l = \text{W}_p^p (\theta_l \sharp \mathcal{F}_x,\theta_l \sharp \hat{\mathcal{F}}_y) $
  \State Compute $w_l = f(\text{W}_p^p (\theta_l \sharp \mathcal{F}_x,\theta_l \sharp \hat{\mathcal{F}}_y))$
  \EndFor
  \State Compute $\widehat{\mathcal{L}_{uniEBSW}} = \left(\frac{1}{L}\sum_{l=1}^L v_l \frac{w_l}{\sum_{i=1}^L w_i}\right)^{\frac{1}{p}}$
  
 \State \textbf{Return:} $\widehat{\mathcal{L}_{uniEBSW}}$
\end{algorithmic}

\end{algorithm*}

\begin{algorithm*}[!t]
\caption{Computational algorithm of the bidirectional EBSW distance}
\label{alg:ISEBSW}
\begin{algorithmic}

\State \textbf{Input:} Features extracted from feature extractor module $\mathcal{F}_x, \mathcal{F}_y$; $p\geq 1$; soft features similarity $\hat{\Pi}$ from Eq.~\ref{eq:soft-similarity}; and the number of projections $L$. \\
  \hrulefill
  \State Compute $\hat{\mathcal{F}}_x = \hat{\Pi}_{yx} \mathcal{F}_x$ and $\hat{\mathcal{F}}_y = \hat{\Pi}_{xy} \mathcal{F}_y$
  \For{$l=1$ to $L$}
  \State Sample $\theta_l \sim \mathcal{U}(\mathbb{S}^{d-1})$
  \State Compute $v_l = \text{W}_p^p (\theta_l \sharp \mathcal{F}_x,\theta_l \sharp \hat{\mathcal{F}}_y) + \text{W}_p^p (\theta_l \sharp \mathcal{F}_y,\theta_l \sharp \hat{\mathcal{F}}_x)$
  \State Compute $w_l = f(\text{W}_p^p (\theta_l \sharp \mathcal{F}_x,\theta_l \sharp \hat{\mathcal{F}}_y) + \text{W}_p^p (\theta_l \sharp \mathcal{F}_y,\theta_l \sharp \hat{\mathcal{F}}_x))$
  \EndFor
  \State Compute $\widehat{\mathcal{L}_{biEBSW}} = \left(\frac{1}{L}\sum_{l=1}^L v_l \frac{w_l}{\sum_{i=1}^L w_i}\right)^{\frac{1}{p}}$
  
 \State \textbf{Return:} $\widehat{\mathcal{L}_{biEBSW}}$
\end{algorithmic}
\end{algorithm*}

\begin{algorithm*}[!t]
\caption{Algorithm of the adaptive refinement}
\label{alg:adaptive-refinement}
\begin{algorithmic}

\State \textbf{Input:} Pair shapes $\mathcal{X}, \mathcal{Y}$ with their Laplace-Beltrami operators $\Phi_x, \Phi_y$ . Trained model with parameter $\mathcal{G}_{\Theta}$. Number of refinement steps $T$. \\

  \hrulefill

  \While{reach T}
  \State Compute $\mathcal{F}_x = \mathcal{G}_{\Theta}(\mathcal{X}, \Phi_x)$ and $\mathcal{F}_y = \mathcal{G}_{\Theta}(\mathcal{Y}, \Phi_y)$. \Comment{Extract features.}
  \State Compute $C_{xy}, C_{yx} = \text{FMSolver}(\mathcal{F}_x, \mathcal{F}_y, \Phi_x, \Phi_y)$. \Comment{Find functional map via FM solver.}
  \State Compute $\tilde{\Pi}_{xy}, \tilde{\Pi}_{yx} = \text{Sinkhorn}(\mathcal{F}_x, \mathcal{F}_y)$. \Comment{Estimate pseudo similarity matrix via Sinkhorn.}
  \State Compute unsupervised losses $\mathcal{L}_{total} (\mathcal{F}_x, \mathcal{F}_y, C_{xy}, C_{yx}, \tilde{\Pi}_{xy}, \tilde{\Pi}_{yx})$. 
  \State Update features and soft similarity matrix by minimizing $\mathcal{L}_{total}$. 
  \EndWhile
  \State Compute $P = NN(\mathcal{F}_x, \mathcal{F}_y)$ \Comment{Compute point-to-point correspondence via nearest neighbor search.}
  
 \State \textbf{Return:} $P$
 
\end{algorithmic}
\end{algorithm*}

\noindent
\textbf{Energy-based sliced Wasserstein distance. }The unidirectional sliced Wasserstein distance version of Eq.~\ref{eq:ISEBSW-align} is defined as: 

\begin{equation}
\begin{aligned}
    \label{eq:uni-ISEBSW-align}
    \mathcal{L}_{uniEBSW} = \left( \frac{\mathbb{E}_{\theta \sim \sigma_0(\theta)} [\text{W}_{\theta, \mathcal{X}} w(\theta) ]}
                            {\mathbb{E}_{\theta \sim \sigma_0(\theta)} [w(\theta)]} \right)^{\frac{1}{p}},
\end{aligned}
\end{equation}
where we denote $\text{W}_{\theta, \mathcal{X}} \coloneqq \text{W}_p^p (\theta \sharp \mathcal{F}_x,\theta \sharp \hat{\mathcal{F}}_y)$, $w(\theta) \coloneqq \frac{\text{exp}(\text{W}_{\theta, \mathcal{X}} )}{\sigma_0(\theta)}$, and $\sigma_0(\theta) \in \mathcal{P}(\mathbb{S}^{d-1})$ denotes the proposed distribution. The unidirectional energy-based sliced Wasserstein distance given in Eq.~\ref{eq:uni-ISEBSW-align} can be computed via importance sampling estimator $L$ Monte Carlo $\theta_1, ..., \theta_L$ sampled from $\sigma_0(\theta)$:

\begin{equation}
\begin{aligned}
    \label{eq:uni-ISEBSW-MCMC}
    \widehat{\mathcal{L}_{uniEBSW}} = \left( \frac{1}{L} \sum_{l=1}^L [\text{W}_{\theta_l, \mathcal{X}} \tilde{w}(\theta_l)] \right)^{\frac{1}{p}},
\end{aligned}
\end{equation}
where $\tilde{w}(\theta_l) \coloneqq \frac{w(\theta_l)}{\sum_{l'=1}^L w(\theta_{l'})}$. When $\sigma_0(\theta) = \mathcal{U}(\mathbb{S}^{d-1}) = \frac{\Gamma(d/2)}{2\pi^{d/2}}$ (a constant of $\theta$)~\cite{nguyen2023energy}, we substitute $w (\theta_l)$ with $f(\text{W}_{\theta_l, \mathcal{X}} )$. We can choose the energy function $f(x) = e^x$, then the normalized importance weights become the Softmax function of $\text{W}_{\theta, \mathcal{X}}$ as follows:
$$\tilde{w}(\theta_l) = \text{Softmax}(\text{W}_{\theta_l, \mathcal{X}}) 
= \frac{\text{exp} (\text{W}_{\theta_l, \mathcal{X}})}{ \sum_{l'=1}^L \text{exp} (\text{W}_{\theta_{l'}, \mathcal{X}})} $$

Based on the computation of unidirectional energy-based sliced Wasserstein distance, we can compute the bidirectional energy-based sliced Wasserstein distance, i.e. $\mathcal{L}_{biEBSW}$, in Eq.~\ref{eq:ISEBSW-align} as follows:

\begin{equation}
\begin{aligned}
    \label{eq:ISEBSW-MCMC}
    \widehat{\mathcal{L}_{biEBSW}} = \left( \frac{1}{L} \sum_{l=1}^L [(\text{W}_{\theta_l, \mathcal{X}} + \text{W}_{\theta_l, \mathcal{Y}}) \hat{w}(\theta_l)] \right)^{\frac{1}{p}},
\end{aligned}
\end{equation}
where we denote $\text{W}_{\theta, \mathcal{Y}} \coloneqq \text{W}_p^p (\theta \sharp \mathcal{F}_y,\theta \sharp \hat{\mathcal{F}}_x)$, and $\hat{w}(\theta_l) \coloneqq \frac{\text{exp} (\text{W}_{\theta_l, \mathcal{X}} + \text{W}_{\theta_l, \mathcal{Y}})}{ \sum_{l'=1}^L \text{exp} (\text{W}_{\theta_{l'}, \mathcal{X}} + \text{W}_{\theta_{l'}, \mathcal{Y}})}$. It is worth noting that the importance weights of $\widehat{\mathcal{L}_{biEBSW}}$ in Eq.~\ref{eq:ISEBSW-MCMC} are different from that of $\widehat{\mathcal{L}_{uniEBSW}}$ in Eq.~\ref{eq:uni-ISEBSW-MCMC}, since the slicing distribution here is shared and affected by both one-dimensional Wasserstein distances, thus providing a more expressive projecting features for computing sliced Wasserstein distance. We provide a pseudo-code for computing the unidirectional and bidirectional energy-based sliced Wasserstein distance in Algorithm~\ref{alg:uni-ISEBSW} and Algorithm~\ref{alg:ISEBSW}, respectively. 

\noindent
\textbf{Adaptive refinement.} As discussed in Sec.~\ref{subsec:adaptive-refinement}, we refine our correspondence result by estimating the pseudo-soft correspondence via entropic regularized optimal transport. The pseudo-code for our adaptive refinement is given in Algorithm~\ref{alg:adaptive-refinement}.

\section{Implementation details}
\label{sec:implementation-details}
All experiments are implemented using Pytorch $2.0$, and executed on a system equipped with an NVIDIA GeForce RTX GPU 2080 Ti and an Intel Xeon(R) Gold 5218 CPU. We employ DiffusionNet~\cite{sharp2020diffusionnet} as the feature extraction mechanism, with wave kernel signatures (WKS)~\cite{aubry2011wave} serving as the input features. The dimension of the WKS is set to $128$ for all of our experiments. Regarding spectral resolution, we opt for the first $200$ eigenfunctions derived from the Laplacian matrices to form the spectral embedding. The output features of the feature extractor are set to $256$. During training, the value of the learning rate is set to $1e-3$ with cosine annealing to the minimum learning rate of $1e-4$. The network is optimized with Adam optimizer with batch size $1$. About adaptive refinement, the number of refinement iterations is empirically set to $12$. 

Regarding the loss functions, as stated in Eq.~\ref{eq:total-losses}, we empirically set $\lambda_1 = \lambda_3 = 1.0, \lambda_2 = 100.0$. About the weight for each component of $\mathcal{L}_{fmap}$ in Eq.~\ref{eq:l_fmap}, we set $\alpha_1 = \alpha_2 = 1.0$. Regarding Sliced Wasserstein distance and energy-based sliced Wasserstein distance, we set $p=2, L=200$ for all of our experiments.

\section{Additional visualizations}
\label{sec:additional-visualizations}
In this section, we provide additional visualizations of our proposed approach on multiple datasets. 

\begin{figure*}[t]
  \begin{center}
    \includegraphics[width=\textwidth]{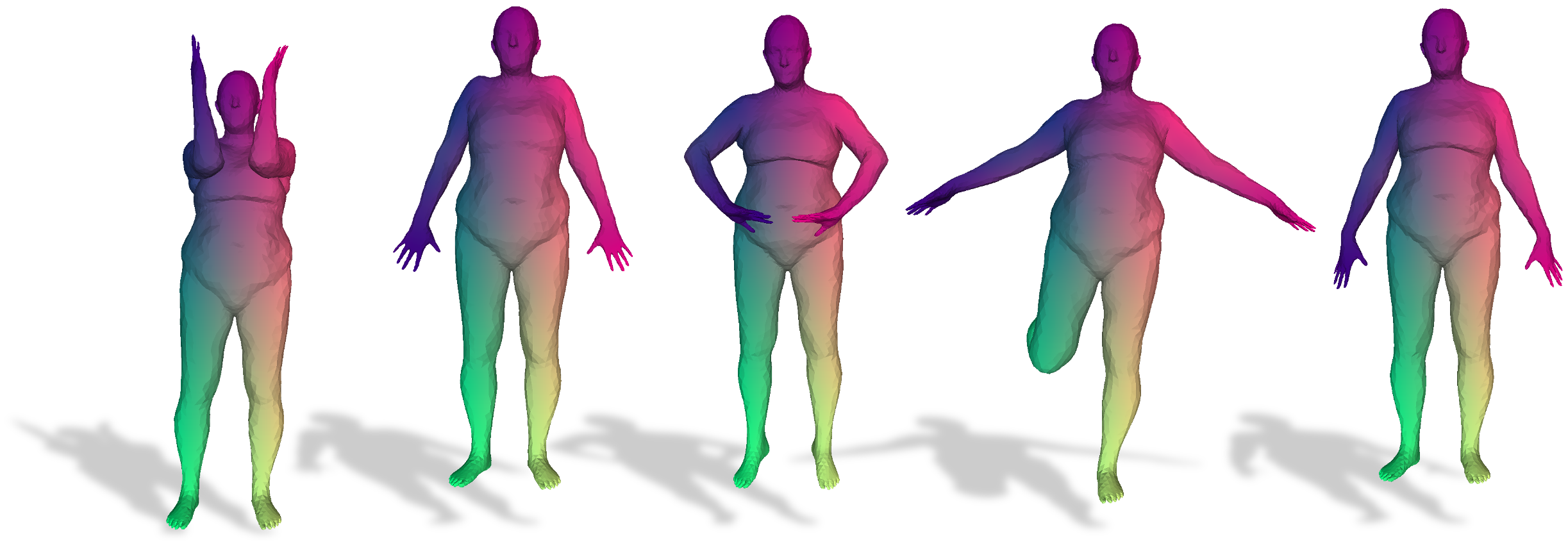}
  \end{center}
  \caption{Qualitative results of our method on FAUST dataset.}
  \label{fig:framework}

\end{figure*}

\begin{figure*}[t]
  \begin{center}
    \includegraphics[width=\textwidth]{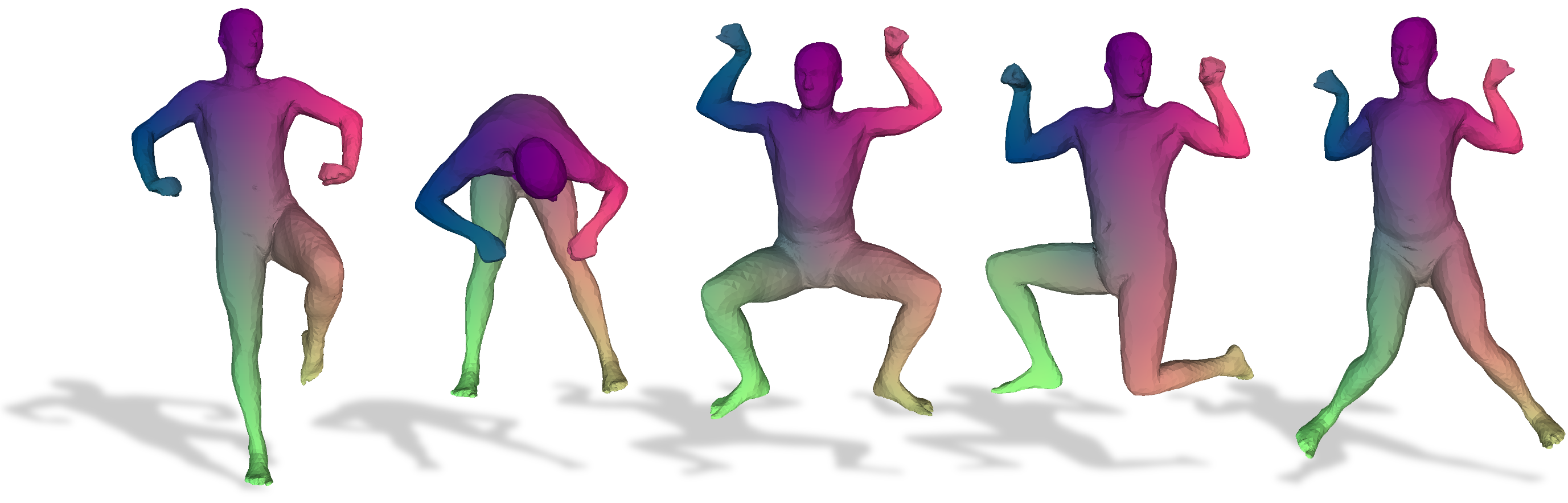}
  \end{center}
  \caption{Qualitative results of our method on SCAPE dataset. }
  \label{fig:framework}

\end{figure*}

\begin{figure*}[t]
  \begin{center}
    \includegraphics[width=\textwidth]{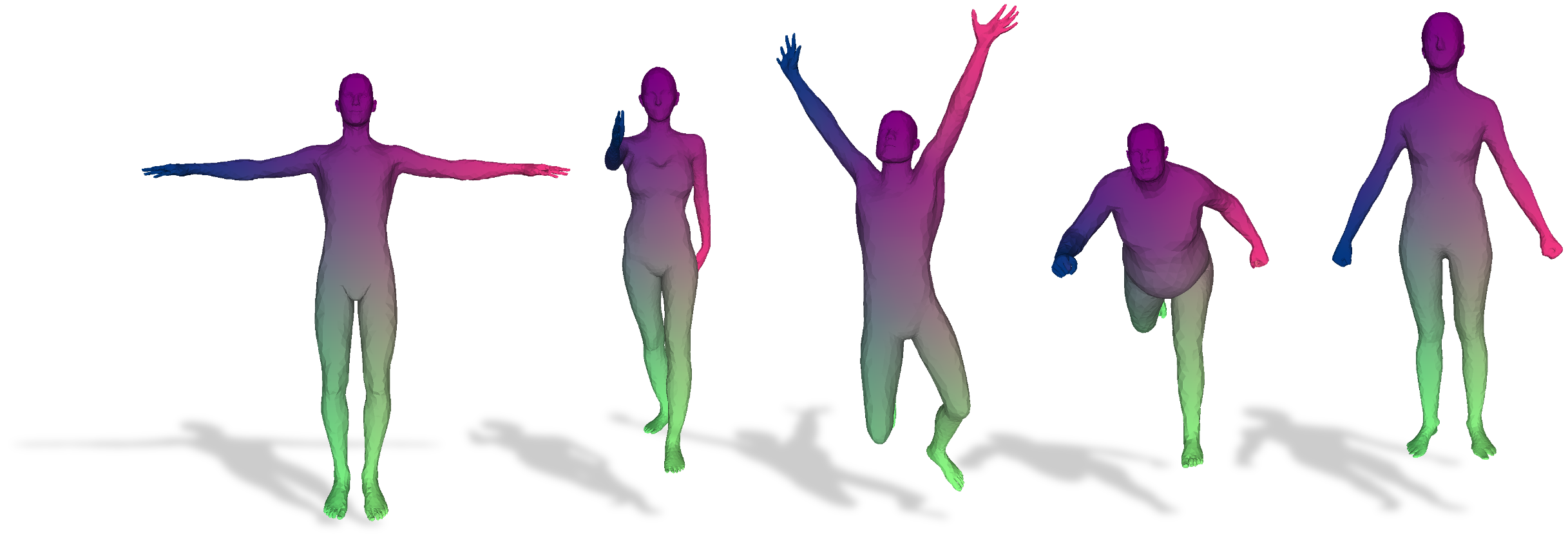}
  \end{center}
  \caption{Qualitative results of our method on SHREC dataset.}
  \label{fig:framework}

\end{figure*}

\begin{figure*}[t]
  \begin{center}
    \includegraphics[width=\textwidth]{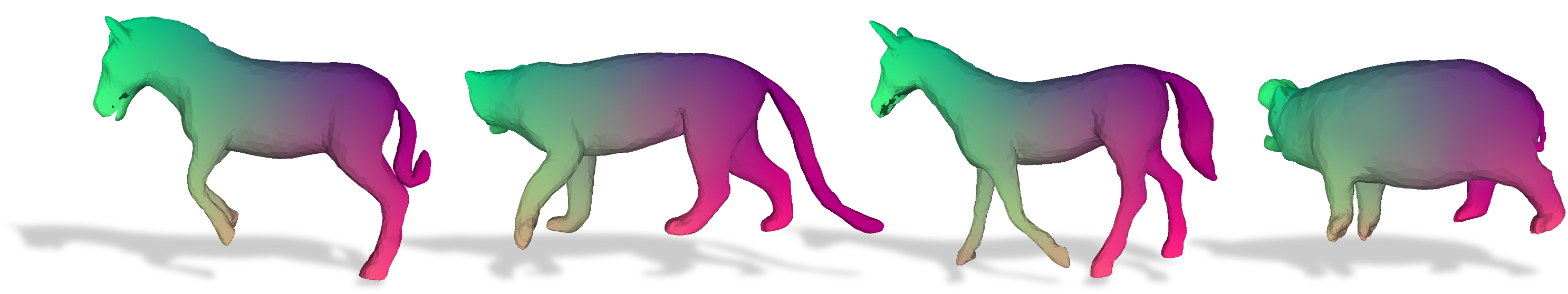}
  \end{center}
  \caption{Qualitative results of our method on SMAL dataset.}
  \label{fig:framework}

\end{figure*}

\begin{figure*}[t]
  \begin{center}
    \includegraphics[width=\textwidth]{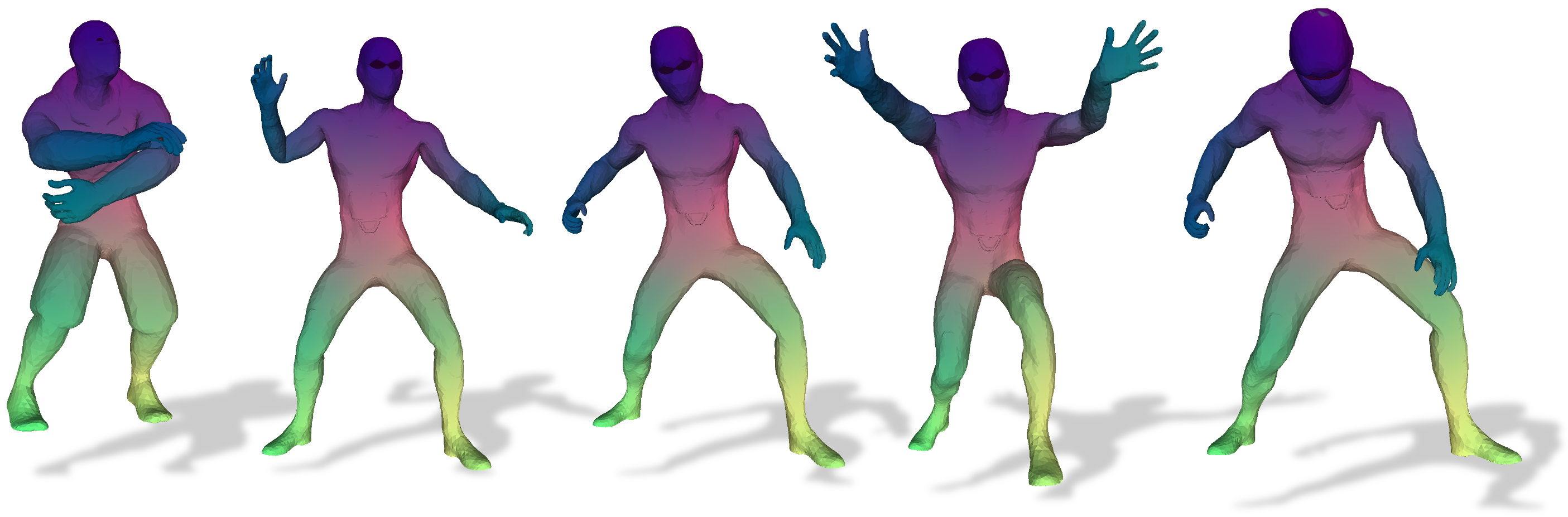}
  \end{center}
  \caption{Qualitative results of our method on DT4D-H dataset.}
  \label{fig:framework}
\end{figure*}

\begin{figure*}[t]
  \begin{center}
    \includegraphics[width=\textwidth]{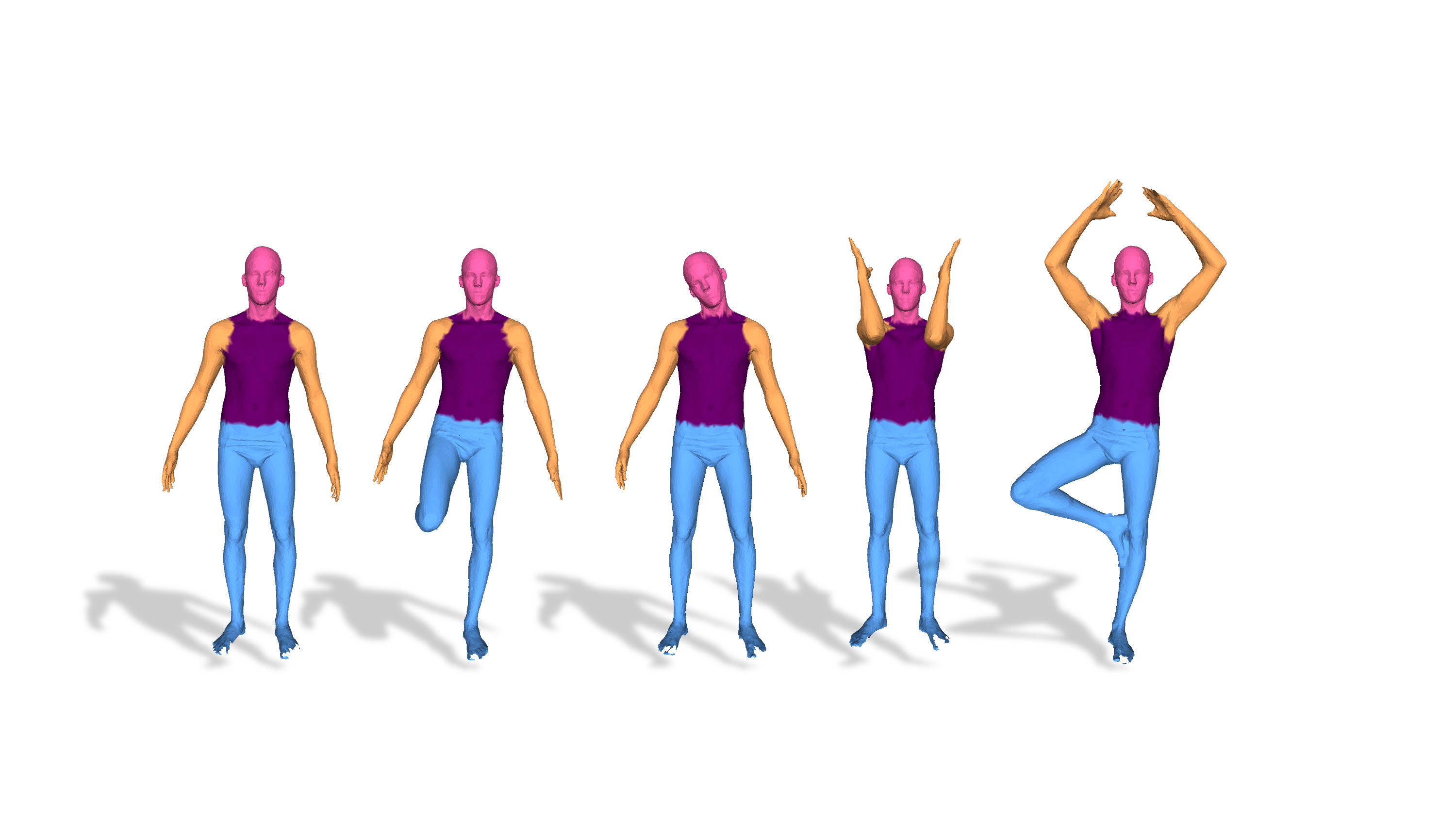}
  \end{center}
  \caption{Qualitative results of our method on segmentation transfer coarse FAUST dataset.}
  \label{fig:framework}
  \vspace{-0.8em}
\end{figure*}

\begin{figure*}[t]
  \begin{center}
    \includegraphics[width=\textwidth]{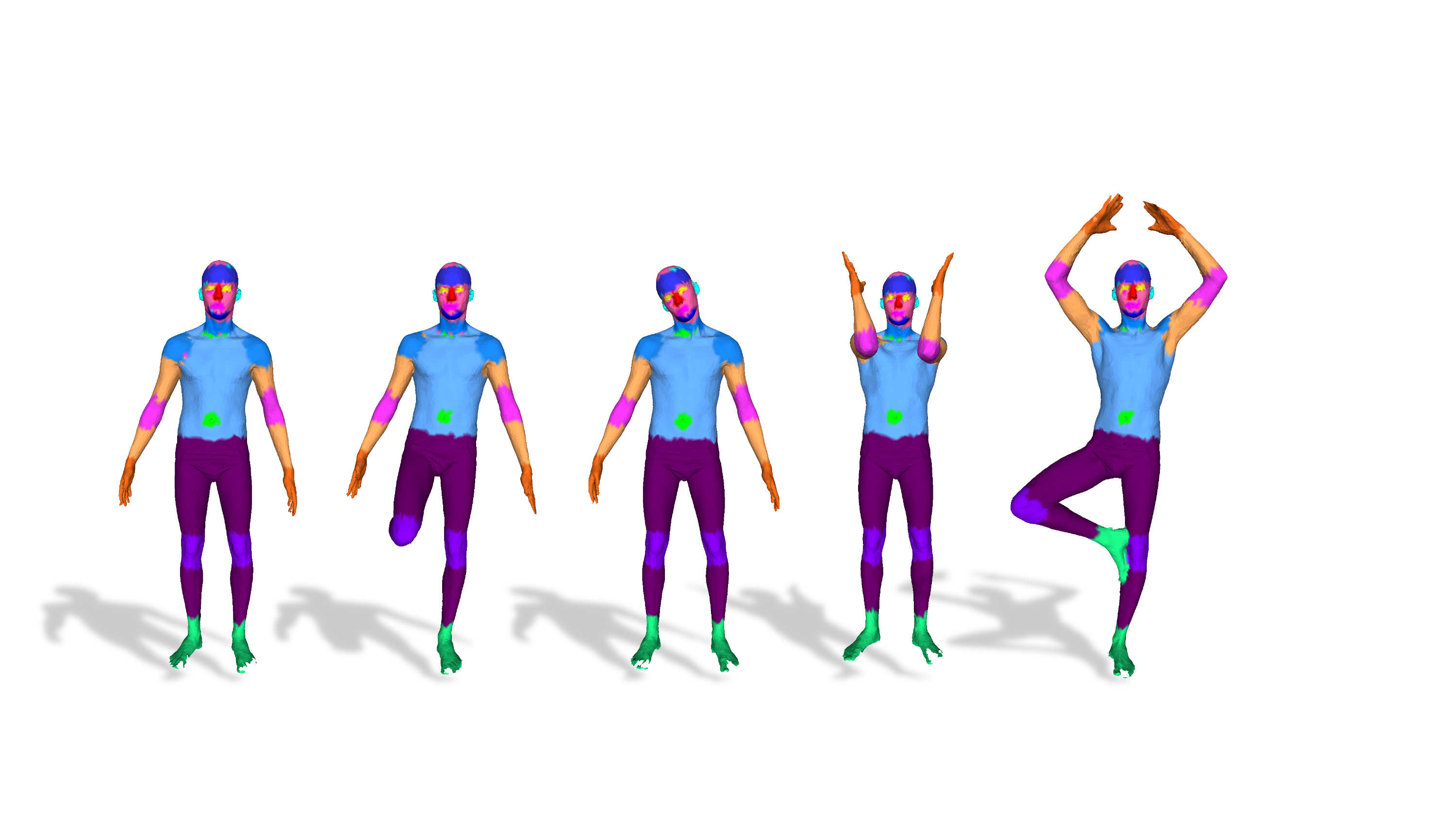}
  \end{center}
  \caption{Qualitative results of our method on segmentation transfer fine-grained FAUST dataset.}
  \label{fig:framework}

\end{figure*}

\end{document}